%% file: main.tex
\documentclass[11pt]{article}
\usepackage{acl}
\usepackage{times}
\usepackage{latexsym}
\usepackage[T1]{fontenc}
\usepackage[utf8]{inputenc}
\usepackage{microtype}
\usepackage{inconsolata}
\usepackage{graphicx}
\usepackage[skins]{tcolorbox}
\usepackage{placeins}
\usepackage{subcaption}
\usepackage{booktabs}
\usepackage{array}
\usepackage{tabularx}
\usepackage{amsmath}
\usepackage{amssymb}
\usepackage{pifont}

\newcommand{\squishlist}{
    \begin{list}{$\bullet$}
    { \setlength{\itemsep}{0pt}
        \setlength{\parsep}{1pt}
        \setlength{\topsep}{1pt}
        \setlength{\partopsep}{0pt}
        \setlength{\leftmargin}{1em}
        \setlength{\labelwidth}{1em}
        \setlength{\labelsep}{0.5em} } }
\newcommand{\squishend}{
    \end{list}  }

\newtcolorbox{insightbox}{
  enhanced,
  colback=teal!20,
  colframe=teal!90!black,
  boxrule=0pt,
  leftrule=5pt,
  rightrule=0pt,
  toprule=0pt,
  bottomrule=0pt,
  sharp corners,
  fontupper=\normalsize,
  before skip=4pt,
  after skip=4pt,
  top=2pt,
  bottom=2pt,
}

\definecolor{langEN}{HTML}{4E79A7}
\definecolor{langLtwo}{HTML}{5E6C84}
\definecolor{langZH}{HTML}{E67E22}
\definecolor{langFR}{HTML}{00ACC1}
\definecolor{langFAS}{HTML}{8E24AA}
\definecolor{langNLD}{HTML}{00897B}
\definecolor{langUKR}{HTML}{F9A825}
\definecolor{langBUL}{HTML}{43A047}
\definecolor{langIND}{HTML}{E53935}
\definecolor{langDEU}{HTML}{6D4C41}
\definecolor{domValues}{HTML}{4E79A7}
\definecolor{domKinship}{HTML}{59A14F}
\definecolor{domReligion}{HTML}{9467BD}
\definecolor{domFood}{HTML}{FF7F0E}
\definecolor{domFestivals}{HTML}{E377C2}
\definecolor{domClothing}{HTML}{00BCD4}
\definecolor{domSymbols}{HTML}{D62728}
\definecolor{domGovernance}{HTML}{7F7F7F}
\definecolor{domIdentity}{HTML}{8C564B}
\definecolor{domDaily}{HTML}{17BECF}
\newcommand{\paperchip}[3]{\begingroup\setlength{\fboxsep}{1.2pt}\colorbox{#1!10}{\textcolor{#1!68!black}{\scriptsize #2\,\textsf{#3}}}\endgroup}
\newcommand{\LangZH}{\paperchip{langZH}{\ding{108}}{ZH}}
\newcommand{\LangFR}{\paperchip{langFR}{\ding{108}}{FR}}
\newcommand{\LangFAS}{\paperchip{langFAS}{\ding{108}}{FAS}}
\newcommand{\LangNLD}{\paperchip{langNLD}{\ding{108}}{NLD}}
\newcommand{\LangUKR}{\paperchip{langUKR}{\ding{108}}{UKR}}
\newcommand{\LangBUL}{\paperchip{langBUL}{\ding{108}}{BUL}}
\newcommand{\LangIND}{\paperchip{langIND}{\ding{108}}{IND}}
\newcommand{\LangDEU}{\paperchip{langDEU}{\ding{108}}{DEU}}
\newcommand{\LangSet}{\LangZH{} \LangFR{} \LangFAS{} \LangNLD{} \LangUKR{} \LangBUL{} \LangIND{} \LangDEU{}}

\newcommand{\coloredlang}[2]{\textcolor{#1!72!black}{\textsc{#2}}}
\newcommand{\coloreddomain}[2]{\textcolor{#1!78!black}{#2}}
\newcommand{\ENONLY}{\coloredlang{langEN}{EN}}
\newcommand{\ENtag}{\coloredlang{langEN}{EN}}
\newcommand{\Ltwotag}{\coloredlang{langLtwo}{L2}}

\newcommand{\ENLtwo}{\ENtag{}+\Ltwotag{}}
\newcommand{\RegimeOne}{\textit{same English exposure}}
\newcommand{\RegimeTwo}{\textit{same total training steps}}

\newcommand{\DomFood}{\coloreddomain{domFood}{\texttt{Food/Cuisine}}}

\newcommand{\DomSymbols}{\coloreddomain{domSymbols}{\texttt{Symbols/Colors}}}
    
\newcommand{\DomIdentity}{\coloreddomain{domIdentity}{\texttt{Social Identity}}}

\definecolor{deltaNeutral}{HTML}{667085}
\definecolor{deltaAdd}{HTML}{1A7F37}
\definecolor{deltaDrop}{HTML}{B42318}

\title{Double Trouble: Bilingual Pretraining Leaves Language-Conditioned Effects in Shared-Language Representations}
\author{
  Anjishnu Mukherjee \
  \textbf{Ziwei Zhu} \
  \textbf{Antonios Anastasopoulos} \\
  George Mason University\\
  \texttt{\{amukher6,zzhu20,antonis\}@gmu.edu}
}

\begin{document}
\maketitle

\begin{abstract}
A concept can carry different associations across languages, while modern language models learn English alongside many other languages during pretraining. Yet comparisons among existing models cannot easily isolate how any one language changes the way these models represent English concepts because their training corpora, compute, architectures, and random seeds all differ. We study this question through a controlled experiment with 40 matched 310M-parameter decoder-only models that share an architecture, tokenizer, training recipe, and English data source. Each bilingual condition adds one of eight languages, while four experimental comparisons separately account for English exposure, total training, and English-document overlap. We align each model pair using 3{,}000 common English words, then measure where 1{,}000 held-out English concepts fall along 50 fixed semantic contrasts, such as red versus white. Across 32 experimental comparisons, English concept positions differ more between bilingual and English-only conditions than between English-only runs with different random seeds. These differences are larger in contextual states than in token embeddings and peak in middle layers. The language learned alongside English can therefore change how a model represents English concepts even when its English input representations are explicitly aligned\footnote{Code available here: \url{https://github.com/iamshnoo/bli}.}
\end{abstract}

\section{Introduction}
\label{sec:introduction}

\begin{figure}[t]
\centering
\includegraphics[width=\columnwidth]{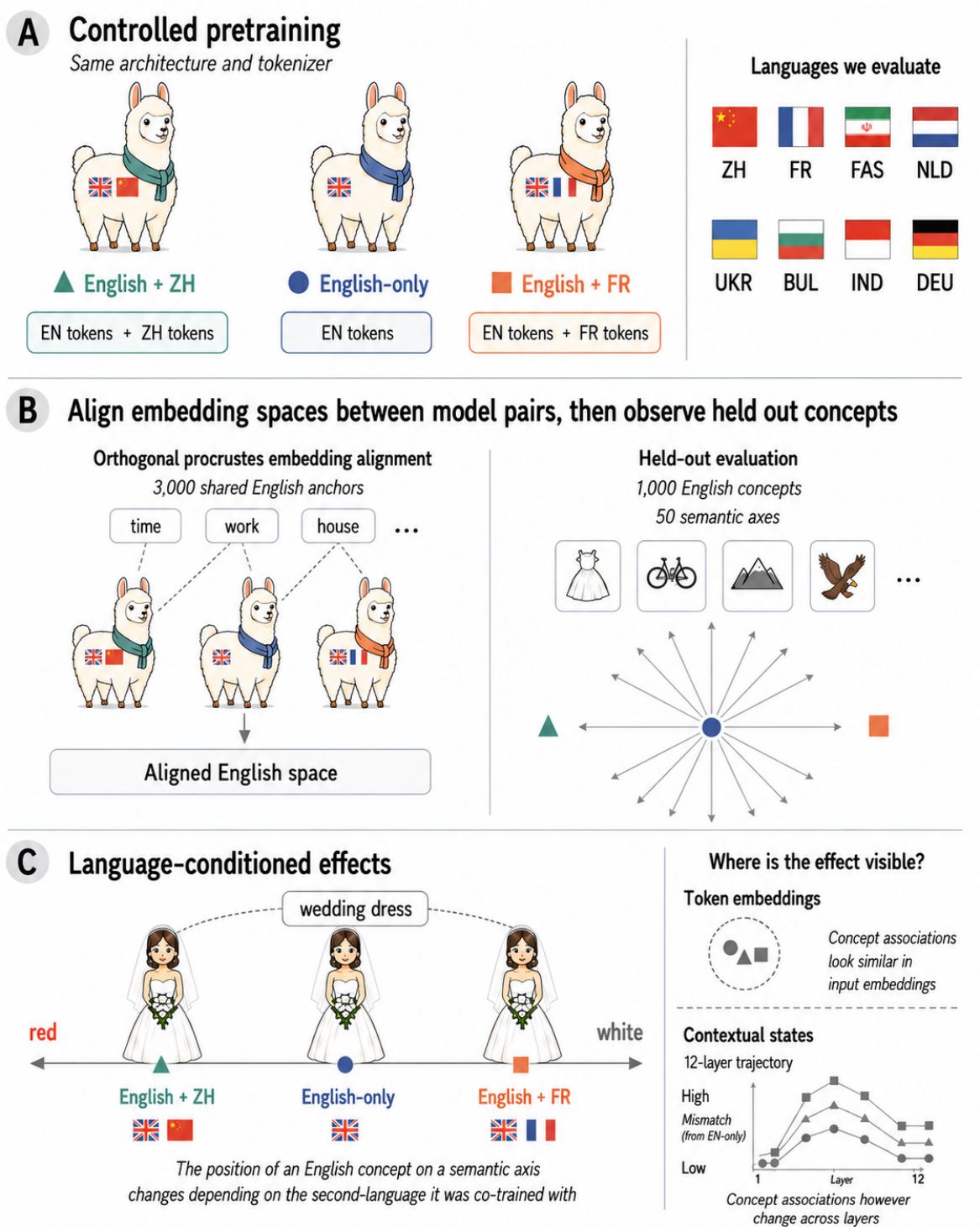}
\caption{\textbf{Aligning English inputs can hide differences in how models represent English concepts.} A bilingual model may place \textit{wedding dress} differently along the red--white contrast than an English-only model. We test 1{,}000 concepts across 50 contrasts and eight additional training languages.}
\label{fig:idea-overview}
\end{figure}

Language models can answer in English while drawing on representations learned from every language in their pretraining mixture. Those languages may change how English concepts are associated with one another. Consider the English concept \textit{wedding dress}. A model trained on English and another language may place it differently along a red--white contrast than a model trained only on English, even after the two models' English token embeddings have been aligned (Figure~\ref{fig:idea-overview}). This example motivates our central question. Does learning another language alongside English change how a model organizes English concepts?

Multilingual pretraining allows one model to transfer across many languages \citep{pires2019multilingual,conneau2020unsupervised}. Shared parameters, however, do not make its representations fully language-neutral. Internal states retain information about language identity and structure \citep{libovicky2020language,chi2020finding}. These studies show that languages remain distinguishable inside multilingual models. They do not tell us whether learning one language changes relations among concepts in another. If it does, aligned English token embeddings may conceal differences in the representations built from them.

Comparisons among existing models cannot isolate this effect. Their corpora, compute budgets, tokenizers, architectures, and random seeds differ, and each difference can alter the learned representations. Linear alignment can remove arbitrary coordinate differences between independently trained embedding spaces \citep{mikolov2013exploiting,smith2017offline,conneau2018word}, but a close fit on the words used to estimate a map does not establish that unseen concepts occupy the same positions. Independent training runs also differ by chance. A convincing comparison must therefore isolate language exposure, reserve new concepts for evaluation, and judge any remaining difference against same-language seed variation.

We construct this comparison with 40 matched 310M-parameter decoder-only models. The bilingual models learn English and one of eight additional languages (\LangZH{} \LangFR{} \LangFAS{} \LangNLD{} \LangUKR{} \LangBUL{} \LangIND{} \LangDEU{}). Every run uses the same architecture, tokenizer, training recipe, and data source. We first compare models that receive the same amount of English, then models that receive the same total number of training steps. We repeat both comparisons using either the same or non-overlapping English documents. English-only runs with different random seeds measure ordinary training variation.

For each model pair, we fit an alignment using 3{,}000 common English words and then test 1{,}000 different English concepts along 50 fixed semantic contrasts. We repeat the comparison on token embeddings and on the contextual states produced at every transformer layer, fitting a new alignment at each layer. The evaluation words never influence the alignment. This design tests whether agreement on common English words extends to new English concepts rather than assuming that it does.

\begin{figure*}[t]
\centering
\includegraphics[width=0.94\textwidth]{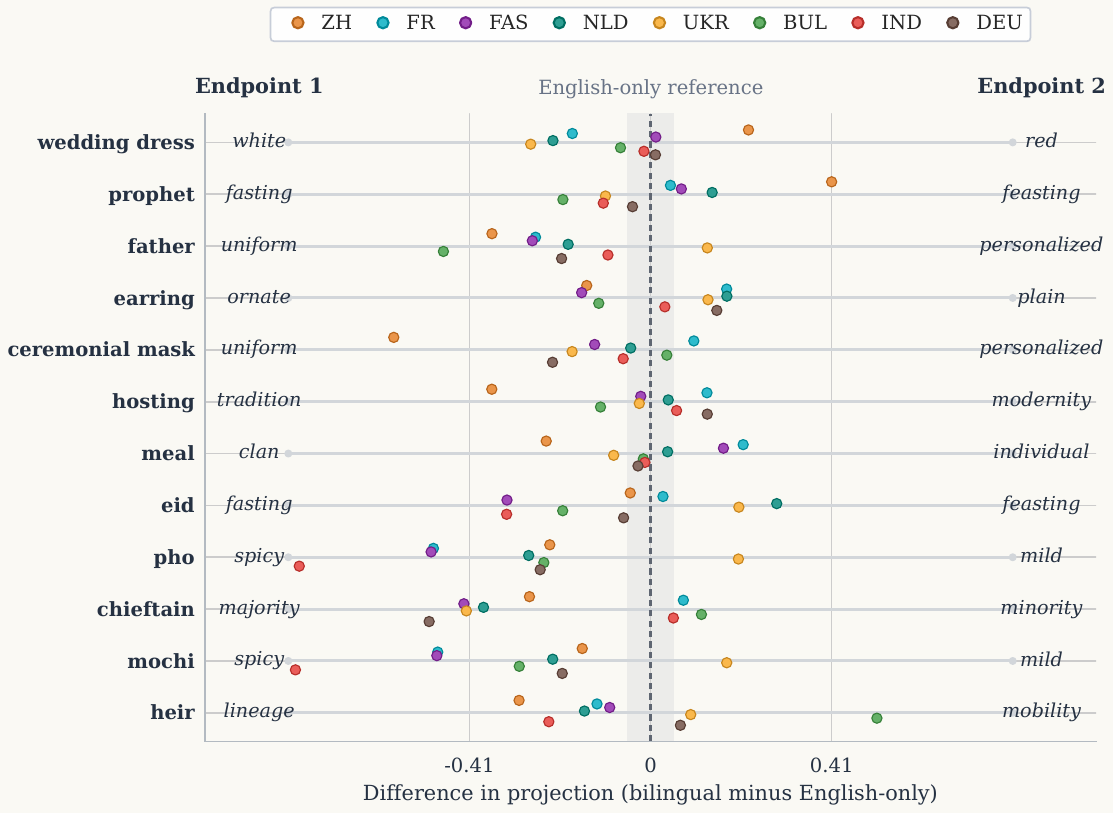}
\caption{\textbf{The same English concept can move toward either end of a semantic contrast depending on the second training language.} Each horizontal line names one contrast, and the colored points show all eight bilingual conditions. The plotted value is the bilingual projection minus the aligned English-only projection. Points to the left move toward the first endpoint, and points to the right move toward the second. These twelve examples show the directional reversals. The aggregate results use all 1{,}000 concepts and 50 contrasts.}
\label{fig:exp4-signed}
\end{figure*}

Across all 32 experimental comparisons, the contextual positions of held-out English concepts differ more between bilingual and English-only models than between English-only runs with different random seeds. These differences are larger than those in token embeddings and, on average, largest in middle layers. Figure~\ref{fig:exp4-signed} shows why direction also matters. The same English concept can move toward opposite ends of a semantic contrast depending on the language learned alongside English. The central result is therefore not merely a gap between embeddings and contextual states. In our controlled models, the additional training language changes where English concepts fall relative to the same semantic contrasts, even after the representations have been explicitly aligned.

The paper makes three contributions.
\squishlist
    \item \textbf{We isolate the effect of learning an additional language.} We train 40 models in which each bilingual model learns English and one of eight other languages. Every model uses the same architecture, tokenizer, data source, and training recipe. The comparisons either match English exposure or total training steps and use shared or non-overlapping English documents. English-only runs with different random seeds establish how much independently trained models normally differ.
    \item \textbf{We evaluate held-out English concepts after alignment.} We align each model pair using 3{,}000 common English words, then measure where 1{,}000 different English concepts fall along 50 fixed semantic contrasts, such as red versus white. We perform this comparison on token embeddings and on contextual states at every transformer layer.
    \item \textbf{We find that the additional language changes how English concepts are related.} Across 32 experimental comparisons, bilingual models place held-out English concepts differently from matched English-only models by more than English-only runs differ from one another. These differences are larger in contextual states than in token embeddings and largest in middle layers on average. Different training languages can also move the same concept toward opposite ends of a contrast.
\squishend

\section{Related Work}
\label{sec:related-work}
Our study connects work on aligning independently trained representations, distinguishing shared from language-specific information, and measuring associations among concepts. Together, these methods let us ask whether an additional training language changes how a model represents English after alignment.

\paragraph{Comparing independently trained representations.} Linear mappings between monolingual embedding spaces are widely used for bilingual lexicon induction, translation, and nearest-neighbor retrieval \citep{schonemann1966generalized,mikolov2013exploiting,smith2017offline,conneau2018word,artetxe2018robust}. Their accuracy depends on the languages, domains, and embedding methods being compared \citep{sogaard2018limitations,vulic2020we,patra2019bilingual}. Related methods support zero-shot translation \citep{mullov-etal-2024-decoupled} and multilingual in-context learning \citep{li-etal-2024-improving-context}. We use alignment to remove arbitrary rotations between two independently trained English representation spaces. The fitted transformation places the spaces in comparable coordinates, but the 1{,}000 concepts used for evaluation are not among the 3{,}000 words used to estimate it. We also fit a separate transformation at every transformer layer. Agreement on the alignment words therefore does not determine the result on the held-out concepts.

\paragraph{Multilingual models.} Pretraining enables cross-lingual transfer \citep{pires2019multilingual,conneau2020unsupervised}. The resulting representations still contain language-specific information \citep{libovicky2020language,chi2020finding}. Prior work has compared multilingual and monolingual representations \citep{zhao-etal-2023-joint}, separated language-sensitive components from shared components \citep{chang-etal-2022-geometry}, recovered typological information \citep{choenni-shutova-2022-investigating}, and traced shared features as they become language-specific outputs \citep{harrasse-etal-2025-tracing}. These studies show that language identity remains visible inside multilingual models. They cannot isolate the effect of adding one language because the models being compared also differ in data, training, or architecture. Our matched models isolate that question. We also vary English-document overlap because repeated documents can change memorization and evaluation results \citep{dodge2021documenting,lee2022deduplicating}. Shared parameters and related writing systems may strengthen cross-lingual transfer \citep{dufter2020identifying,muller2021unseen}, but our question is whether the added language also changes relations among English concepts.

\paragraph{Interpreting concept associations.} Directions between contrasting words have been used to study bias and historical change in word embeddings \citep{bolukbasi2016man,caliskan2017semantics,garg2018word}. Related cross-cultural work examines value associations in pretrained models \citep{arora2023probing}. We use each contrast, such as \textit{individual} versus \textit{collective}, as a fixed reference for comparing the same English concept across training conditions. These contrasts do not assign a score to a language, culture, or speaker. Research on cross-cultural values motivated the nine broad themes \citep{hofstede2001culture,schwartz2006theory,inglehart2005modernization,house2004culture}, but we wrote the endpoint pairs and fixed them before analysis. We retain the direction of each change because an average absolute difference would hide cases in which different training languages move the same concept toward opposite endpoints.

\section{Experimental Design}
\label{sec:study-design}

Each experiment compares an English-only model with a bilingual model after aligning their English representations. The models share an architecture, tokenizer, training recipe, and data source. Separate comparisons account for English exposure, total training, English-document overlap, and random initialization. Held-out concepts test whether the alignment extends beyond the words used to estimate it.

\subsection{What We Measure}
\label{sec:target-phenomenon}

For each English-only model \(M_{\text{\ENtag{}}}\) and matched bilingual model \(M_{\text{\ENLtwo{}}}\), we measure disagreement on held-out English concepts after alignment. Pairs of English-only models trained with different random seeds provide the reference. For any disagreement measure \(m\), we report
\[
\Delta m \;=\; m(M_{\text{\ENtag{}}}, M_{\text{\ENLtwo{}}})\;-\;\bar m_{\text{seed-var}},
\]
where \(\bar m_{\text{seed-var}}\) is the average disagreement between English-only runs that differ only in random seed. A value near zero means that the bilingual and English-only models differ by about as much as two English-only runs. Positive values indicate disagreement beyond this same-language reference.

\subsection{Training and Controls}
\label{sec:training}

\paragraph{Data and languages.}
BabyBabelLM provides comparable 100-million-token collections for eight non-English languages \citep{jumelet2026babybabellm}. We use Chinese, French, Persian, Dutch, Ukrainian, Bulgarian, Indonesian, and German (\LangZH{} \LangFR{} \LangFAS{} \LangNLD{} \LangUKR{} \LangBUL{} \LangIND{} \LangDEU{}). Every run uses the same 310M-parameter decoder-only architecture and training recipe. Under this recipe, 1{,}500 steps process about 50 million tokens and 3{,}000 steps process about 100 million tokens. The appendix gives the complete architecture, batching, and tokenizer details.

\paragraph{What we hold fixed.}
Three training differences could otherwise explain the result. A bilingual model may see less English, receive more training, or share documents with its English-only counterpart. We test each explanation directly.
\squishlist
    \item \textbf{\RegimeOne{}.} The English-only model trains for 1{,}500 steps. The bilingual model trains for 3{,}000 steps, alternating equal blocks of English and the additional language. Both models therefore receive exactly 1{,}500 English steps.
    \item \textbf{\RegimeTwo{}.} Both models train for 3{,}000 steps. The English-only model receives only English, while the bilingual model receives 1{,}500 English steps and 1{,}500 steps in the additional language.
\squishend
We repeat both training comparisons under two document conditions.
\squishlist
\item \textbf{Shared English documents.} Both models train on the same English documents.
\item \textbf{Separate English documents.} The models train on non-overlapping English documents.
\squishend
Table~\ref{tab:control-matrix} shows the resulting four comparisons and the explanation tested by each one.

\begin{table*}[t]
\centering
\footnotesize
\setlength{\tabcolsep}{3.2pt}
\renewcommand{\arraystretch}{1.16}
\begin{tabular}{@{}>{\raggedright\arraybackslash}p{0.16\textwidth}>{\raggedright\arraybackslash}p{0.25\textwidth}>{\centering\arraybackslash}p{0.13\textwidth}>{\centering\arraybackslash}p{0.13\textwidth}>{\centering\arraybackslash}p{0.14\textwidth}>{\centering\arraybackslash}p{0.10\textwidth}@{}}
\toprule
\textbf{Comparison} &
\textbf{Question answered} &
\textbf{English steps} &
\textbf{Total training steps} &
\textbf{English documents} &
\textbf{Median contextual} \(\Delta D_{Axis}\) \\
\midrule
Same English exposure &
Does a difference remain when both models see the same amount of English? &
\((1500,1500)\) &
\((1500,3000)\) &
shared &
\textbf{0.16} \\
\midrule
Same English exposure, separate documents &
Does the difference remain with no English documents in common? &
\((1500,1500)\) &
\((1500,3000)\) &
non-overlapping &
\textbf{0.18} \\
\midrule
Same total training steps &
Does a difference remain after matching the number of training steps? &
\((3000,1500)\) &
\((3000,3000)\) &
shared &
\textbf{0.53} \\
\midrule
Same total training steps, separate documents &
Does the difference remain without shared English documents? &
\((3000,1500)\) &
\((3000,3000)\) &
non-overlapping &
\textbf{0.32} \\
\bottomrule
\end{tabular}
\caption{\textbf{The four comparisons test English exposure, total training, and English-document overlap.} Each numerical pair lists the English-only model first and the bilingual model second. Every bilingual model receives 1{,}500 English steps and 1{,}500 steps in the additional language. The English-only model receives either 1{,}500 steps to match English exposure or 3{,}000 steps to match total training. The median contextual difference exceeds English-only seed variation in all four comparisons.}
\label{tab:control-matrix}
\end{table*}

The same-English-exposure comparison with separate documents is the strictest control for English data. Both models receive 1{,}500 English steps drawn from non-overlapping documents. Only the bilingual model receives another 1{,}500 steps in the additional language. The matched-total-training comparisons address a different explanation by giving both models the same total number of steps.

\subsection{Evaluation Concepts and Semantic Contrasts}
\label{sec:probes}

We estimate each alignment using 3{,}000 common English words selected by frequency and filtered to remove stop words, non-ASCII text, and overlap with the evaluation sets. A separate set of 1{,}000 English concepts covers values and norms, family and kinship, religion and ritual, food and cuisine, festivals and holidays, clothing and appearance, symbols and colors, governance and law, social identity, and daily customs.

Each of the 50 semantic contrasts pairs two endpoints, such as \textit{individual} and \textit{collective}. The contrasts span nine themes motivated by research on cross-cultural values \citep{hofstede2001culture,schwartz2006theory,inglehart2005modernization,house2004culture} and prior cross-cultural NLP analysis \citep{arora2023probing}. We wrote the endpoint pairs ourselves and fixed them before examining the results. They provide common reference directions for comparing concepts, not scores assigned to a language, culture, or speaker. Table~\ref{tab:main-axis-inventory} gives one example from each theme.

We also evaluate 100 common physical terms, including body parts, adapted from basic-vocabulary traditions \citep{swadesh1955towards}. These negative controls test whether the difference is limited to the primary concept categories. For the non-English extension, we translate the words with NLLB \citep{costa2022no}, score translation quality with COMETKiwi \citep{rei2022cometkiwi}, back-translate them, and manually review low-confidence cases. The primary analysis uses only the original English words. Complete word lists and translation checks appear in Appendix~\ref{sec:probe-details}.

\subsection{Alignment and Metrics}
\label{sec:metrics}

For each model pair $(A,B)$, token embeddings are compared only with token embeddings, and contextual states are compared only with contextual states. Let $X^a, X^b \in \mathbb{R}^{V\times d}$ denote either representation. We use the English alignment words $\mathcal{N}$ to fit an orthogonal Procrustes transformation.
\[
W^\star=\arg\min_{W^\top W=I}\|X^a_{\mathcal N}W-X^b_{\mathcal N}\|_F.
\]
This transformation makes the coordinate systems comparable without changing distances or angles within either space. We then measure disagreement on the held-out concepts $\mathcal C$. Our primary measure, \textbf{axis projection disagreement}, asks whether the same concept occupies the same relative position along a semantic contrast after alignment.
\[
D_{Axis,i}=\frac{1}{|\mathcal C|}
\sum_{w\in\mathcal C}
\left|\left\langle x_w^a,\hat u_i^a\right\rangle-\left\langle x_w^b,\hat u_i^b\right\rangle\right|,
\]
for contrast $i$. After applying the fitted transformation to model $A$, $x_w^a$ and $x_w^b$ are the aligned representations of concept $w$. The unit vectors $\hat u_i^a$ and $\hat u_i^b$ point from the first endpoint of the contrast to the second in each model. The notation $\langle\cdot,\cdot\rangle$ denotes the Euclidean inner product.

\(D_{Axis}\) is the absolute difference between the two projections. For examples such as Figure~\ref{fig:exp4-signed}, we retain the sign to show which endpoint the bilingual representation moves toward relative to the English-only representation. Two complementary measures compare each concept's 25 nearest neighbors, denoted by \(D_{NN}\), and the pairwise cosine similarities among concepts, denoted by \(D_{Struct}\). Experiment~4 also tests a less constrained affine transformation.

Every reported \(\Delta\) subtracts the English-only seed mean defined above. Positive values therefore exceed the average disagreement between English-only runs. At each checkpoint, this reference uses six ordered comparisons corresponding to three unique seed pairs.

\section{Results}
\label{sec:experiments}

Unequal English exposure, longer training, or reused documents could each create an apparent language effect. Experiments 1 and 2 test these alternatives. Experiments 3 and 4 determine where the remaining difference appears across layers and checkpoints and whether it survives other alignment methods.

\subsection{Experiment 1. Does the difference remain after matching English exposure or total training?}
\label{sec:exp1}

If either unequal English exposure or unequal total training causes the difference, matching the relevant quantity should reduce it to ordinary seed variation. The first comparison gives both models the same number of English steps. The second gives both models the same total number of training steps.

\begin{figure}[!t]
\centering
\includegraphics[width=\columnwidth]{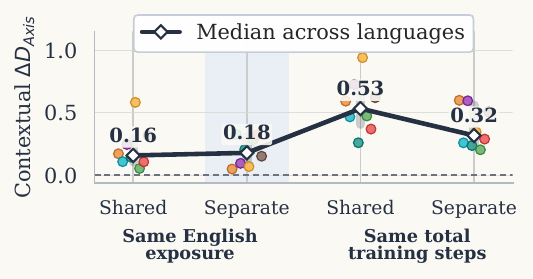}
\caption{\textbf{The contextual difference exceeds English-only seed variation in all four comparisons.} Each point represents one additional language, and the dark line shows the median. The four positions match either English exposure or total training and use shared or separate English documents. Zero is the average difference between English-only runs.}
\label{fig:exp1-controls}
\end{figure}

Figure~\ref{fig:exp1-controls} shows all eight additional languages under the four comparisons in Table~\ref{tab:control-matrix}. Every contextual value exceeds the English-only seed mean, although some language-level intervals are wide. The difference remains when the paired models receive the same amount of English and train on the same English documents. Less English and different English text are therefore not sufficient explanations.

\begin{insightbox}
\textbf{Takeaway 1.} Contextual differences remain above English-only seed variation when either English exposure or total training is matched.
\end{insightbox}

\subsection{Experiment 2. Is the difference driven by shared English documents?}
\label{sec:exp2}

\begin{figure}[t]
\centering
\includegraphics[width=\columnwidth]{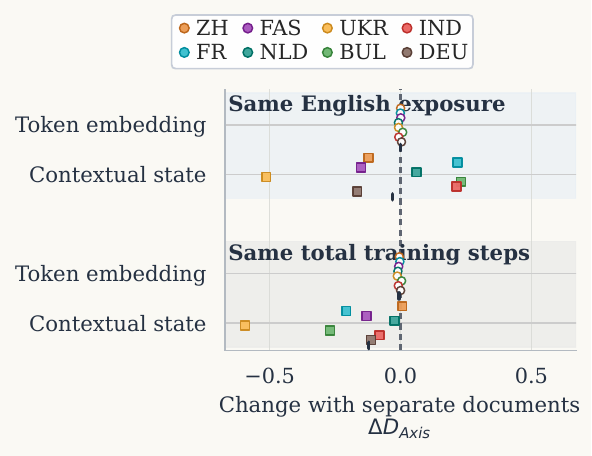}
\caption{\textbf{Removing shared English documents does not consistently reduce the contextual difference.} Each point shows the change after replacing shared English documents with non-overlapping sets. Token-embedding changes remain near zero, while contextual changes vary in direction across languages.}
\label{fig:exp2-overlap}
\end{figure}

Training on the same English documents could make two models appear similar for reasons unrelated to language exposure. We therefore repeat every comparison with non-overlapping English documents. Removing shared documents does not consistently increase or decrease the contextual difference across languages (Figure~\ref{fig:exp2-overlap}). Changes in token embeddings remain close to zero. The contextual difference still exceeds English-only seed variation for most language conditions, with language-level uncertainty reported in Appendix Figure~\ref{fig:appendix-ci-guide}.

\begin{insightbox}
\textbf{Takeaway 2.} Sharing English documents does not account for the difference between bilingual and English-only models.
\end{insightbox}

\subsection{Experiment 3. Where in the model is the difference largest?}
\label{sec:exp3}

Aligning token embeddings constrains only the model input. Each transformer layer can still change how a concept is represented in context. We therefore fit a separate alignment to the contextual states from each of the 12 layers. We also normalize every vector to unit length and standardize projections using neutral English words. These two checks test whether vector length alone produces the layerwise pattern.

\begin{figure*}[t]
\centering
\begin{subfigure}[t]{0.49\textwidth}
\centering
\includegraphics[width=0.96\linewidth]{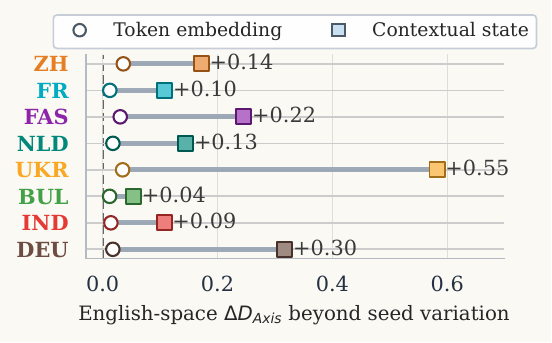}
\caption{In English, contextual states differ more than token embeddings for every additional training language.}
\label{fig:exp4-ratio}
\end{subfigure}
\hfill
\begin{subfigure}[t]{0.49\textwidth}
\centering
\includegraphics[width=0.96\linewidth]{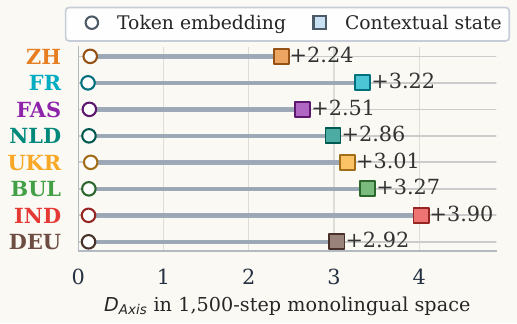}
\caption{The same ordering holds when evaluation is performed in each additional language.}
\label{fig:multilingual-combined}
\end{subfigure}
\caption{\textbf{Contextual states differ more than token embeddings in both English and the additional language.} \textbf{(a)} In English, squares show contextual states and circles show token embeddings for each additional training language. \textbf{(b)} Evaluating concepts in each additional language yields the same ordering.
}
\label{fig:representation-gap-panel}
\end{figure*}

For every additional language, contextual states differ more than token embeddings when the concepts are evaluated in English (Figure~\ref{fig:exp4-ratio}). The same ordering holds when concepts are evaluated in the additional language (Figure~\ref{fig:multilingual-combined}). The signed examples in Figure~\ref{fig:exp4-signed} show that the aggregate difference can arise from movement in either direction along a contrast.

\begin{figure*}[t]
\centering
\begin{subfigure}[t]{0.49\textwidth}
\centering
\includegraphics[width=0.98\linewidth,height=0.245\textheight,keepaspectratio]{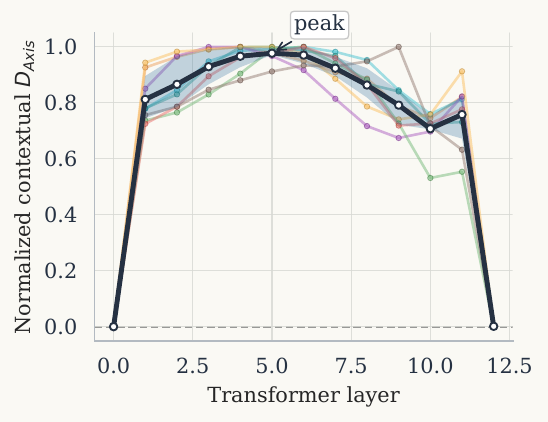}
\caption{Difference across transformer layers.}
\label{fig:exp4-layerwise}
\end{subfigure}
\hfill
\begin{subfigure}[t]{0.49\textwidth}
\centering
\includegraphics[width=0.98\linewidth,height=0.245\textheight,keepaspectratio]{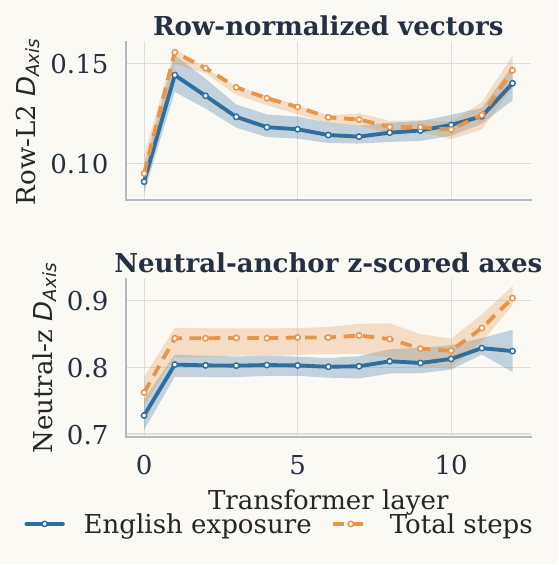}
\caption{Differences across layers after controlling for vector length.}
\label{fig:norm-layerwise}
\end{subfigure}
\caption{\textbf{Differences in English concept placement are largest in middle layers on average.} \textbf{(a)} Thin lines show the eight additional languages, the dark line shows their mean, and the band spans the 20th to 80th percentiles. \textbf{(b)} Normalizing vector lengths and standardizing against neutral English words preserve the middle-layer peak.}
\label{fig:layer-norm-panel}
\end{figure*}

\begin{figure}[!b]
\centering
\includegraphics[width=0.84\columnwidth]{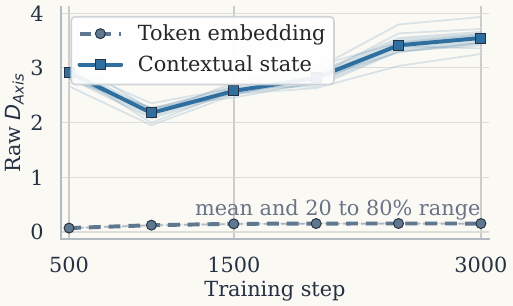}
\caption{\textbf{Differences in English concept placement appear at the first saved checkpoint and persist through training.} Lines average the eight additional languages under matched English exposure with shared English documents. Checkpoints are 500 steps apart. Bands span the 20th to 80th percentiles across languages.}
\label{fig:dense-progress-main}
\end{figure}

The average difference is small at the aligned input layer, rises through the middle transformer layers, and decreases in the final layers while remaining above its input value (Figure~\ref{fig:exp4-layerwise}). Because we fit a new alignment at every layer, this rise cannot be attributed only to using an input-layer transformation on later states. Normalizing vector lengths or standardizing projections against neutral English words preserves the middle-layer peak (Figure~\ref{fig:norm-layerwise}). Vector length alone therefore does not explain the result.

\begin{insightbox}
\textbf{Takeaway 3.} English concept positions differ most in middle layers on average, even after controlling for vector length.
\end{insightbox}

\begin{table*}[t]
\centering
\footnotesize
\setlength{\tabcolsep}{3pt}
\renewcommand{\arraystretch}{1.18}
\input{tables/tab17-exp4-alignment-methods.tex}
\caption{\textbf{The difference remains when alignment is fitted directly to contextual states.} Across sixteen shared-document comparisons, contextual \(\Delta D_{Axis}\) is 0.386 under either orthogonal fit and 0.580 with an affine transformation. English-only seed variation is subtracted, and every concept is held out from alignment.}
\label{tab:exp5-alignment}
\end{table*}

\begin{table*}[t]
\begin{minipage}[t]{0.48\textwidth}
\centering
\footnotesize
\setlength{\tabcolsep}{2.2pt}
\renewcommand{\arraystretch}{1.08}
\input{tables/tab-main-axis-inventory.tex}
\captionof{table}{\textbf{The 50 fixed semantic contrasts cover nine themes.} One example and the number of contrasts in each theme are shown. Appendix Tables~\ref{tab:appendix-axes} and~\ref{tab:appendix-axis-grounding} give the full list and sources.}
\label{tab:main-axis-inventory}
\end{minipage}\hfill
\begin{minipage}[t]{0.49\textwidth}
\centering
\footnotesize
\setlength{\tabcolsep}{2.0pt}
\renewcommand{\arraystretch}{1.08}
\input{tables/tab-main-translation-quality.tex}
\captionof{table}{\textbf{Translation quality varies, but the document-overlap result is similar across quality groups.} The final column gives the largest absolute median change in contextual neighbor overlap after using non-overlapping English documents.}
\label{tab:main-translation-quality}
\end{minipage}
\end{table*}

\subsection{Experiment 4. Does the difference appear early and remain under other alignments?}
\label{sec:exp4}

If the difference emerges only at the end of training, it should be absent from earlier checkpoints. We therefore evaluate every saved checkpoint at intervals of 500 steps. We also fit the orthogonal transformation directly to contextual states and test a less constrained affine transformation. If the original alignment creates the difference, these alternatives should substantially reduce it. Every transformation is evaluated on concepts that were not used for alignment (Table~\ref{tab:exp5-alignment}).

The contextual difference appears at the first 500-step checkpoint and remains at every saved checkpoint through 3{,}000 steps (Figure~\ref{fig:dense-progress-main}). On held-out concepts, mean \(\Delta D_{Axis}\) is 0.386 whether the orthogonal transformation is fitted to token embeddings or contextual states. The affine transformation yields 0.580 (Table~\ref{tab:exp5-alignment}). Changing the subset of English words used for alignment also preserves the separation between token embeddings and contextual states (Appendix Table~\ref{tab:anchor-sensitivity}). The result is therefore neither confined to the final checkpoint nor removed by a more flexible alignment.

\begin{insightbox}
\textbf{Takeaway 4.} The difference appears early and remains when the orthogonal transformation is fitted directly to contextual states or replaced with an affine transformation.
\end{insightbox}

\section{Discussion}
\label{sec:discussion}

\paragraph{What alignment shows.} Aligning common English words shows that two independently trained models can express those words in comparable coordinates. It does not show that other concepts retain the same relationships. The representations of our held-out concepts remain different even when the transformation is fitted directly to contextual states. Prior work has also found that multilingual representations change across layers \citep{chi2020finding,wendler2024llamas}. A layer-level comparison should therefore fit the alignment at that layer, evaluate different concepts, and compare the remaining disagreement with same-language training variation.

\paragraph{How to interpret the directional changes.} Figure~\ref{fig:exp4-signed} places all eight language conditions between explicitly named endpoints. The same concept can move toward either endpoint depending on the additional training language. A single ranking would hide this result. The absolute measure summarizes how far the models disagree, while the signed examples show the direction of individual changes. The contrasts listed in Table~\ref{tab:main-axis-inventory} are fixed references for concepts, not scales for ranking languages or cultures.

\paragraph{What the training controls rule out.} The four comparisons address different explanations. Matching English exposure shows that the result is not explained only by the bilingual model seeing less English. Matching total training shows that it is not explained only by additional training steps. Using non-overlapping English documents shows that repeated English text is not required. These controls do not identify the mechanism. Cross-language differences in word meaning, co-occurrence, or the way related meanings compete for the same model parameters may all contribute. The middle-layer result is consistent with contextual processing strengthening the difference, but it does not distinguish among these explanations.

\paragraph{Where translation quality matters.} The main analysis uses the original English concepts and does not depend on translation. Translation is required only when we repeat the evaluation in the additional language (Figure~\ref{fig:multilingual-combined}). Mean COMETKiwi scores range from 0.745 to 0.797, and the proportion scoring at least 0.80 ranges from 52.9\% to 71.7\% (Table~\ref{tab:main-translation-quality}). Across these quality groups, replacing shared English documents with separate sets changes median contextual neighbor overlap by at most 0.028. Translation quality is therefore unlikely to explain the document-overlap result, although individual translations still warrant further review.

\paragraph{Connection to prior work.} \citet{wendler2024llamas} found that Llama-2's intermediate states become more similar to English before later states move toward the language of the input. That work explains how several languages coexist within one multilingual model. We ask a complementary question across matched models. What changes within English when another language is added to training? Our results show that the added language changes measurable relations among English concepts. Multilingual training can therefore retain language-specific information while also changing representations within English.

\paragraph{Implications for model comparison.} Two models can align closely on 3{,}000 common English words while disagreeing on 1{,}000 different concepts. An alignment score on its own is therefore insufficient evidence that the models represent English similarly. The comparison should be made at the layer being interpreted, using concepts that did not determine the alignment. Reporting both direction and magnitude distinguishes opposing changes from a common shift, and English-only runs reveal whether the disagreement exceeds ordinary training variation.

\paragraph{Open questions.} Our 310M-parameter decoder-only models provide a controlled first measurement at one scale. The same held-out comparison should be repeated with larger architectures, mixtures containing several additional languages, and multiple seeds for every bilingual condition. More languages are also needed before testing whether script or language family predicts the direction or size of a change. Finally, output evaluations must determine when a difference in internal representations changes generated text.

\section{Conclusion}
\label{sec:conclusion}

Agreement on familiar English words does not imply agreement on new concepts. In our controlled comparisons, an additional training language changed where held-out English concepts fell along fixed semantic contrasts. The difference exceeded English-only seed variation, varied in direction across languages, and peaked in middle layers on average. Comparisons should therefore fit and evaluate alignments on separate concepts at the relevant layer and use same-language runs as a reference. More broadly, a pretraining mixture can shape relations within English even when English input representations appear aligned.

\section*{Limitations}
\label{sec:limitations}

We study one 310M-parameter decoder-only architecture so that training conditions can be closely matched. The evidence therefore does not establish that the same pattern holds at larger scales or in other architectures. Each bilingual condition uses one random seed. English-only runs estimate ordinary seed variation, but repeated bilingual runs are needed to measure uncertainty within each language condition. The evaluation also depends on 50 hand-authored semantic contrasts and, for the non-English extension, translated concepts. These contrasts provide fixed reference directions rather than a complete account of any language or culture. We measure internal representations, not generated behavior, so the results do not show that the observed differences change model outputs.

\section*{Ethical Considerations}
\label{sec:ethical}

This study analyzes model representations and neither deploys a system nor uses personal or user-generated data. The main risk lies in how the results are interpreted. A difference associated with one training language must not be treated as a property of that language, its cultures, or its speakers. Every language contains substantial regional, social, and historical variation, while our training data and hand-authored contrasts capture only narrow samples. The contrasts are tools for comparing models, not scales for ranking languages or people.

We report every evaluated language rather than selecting only large effects. We retain the direction of individual changes, compare them with English-only seed variation, and include common physical terms as negative controls. We also distinguish the primary English analysis from the translated extension and report translation-quality checks. These choices make the scope of the evidence clearer, but they do not make the concepts culturally comprehensive or guarantee every translation.

The results concern internal representations, not generated behavior. They should not be used to infer a speaker's identity, evaluate the quality of a language, or justify removing a language from training. Instead, they motivate direct evaluation of model outputs. Before drawing conclusions about deployment, developers should test whether differences among internal English concept representations produce corresponding differences in generated text.

\section*{Acknowledgments}
\label{sec:ack}
We are thankful to the reviewers who provided feedback in earlier versions of this work. This work was generously supported by the US National Science Foundation CAREER award 2439202. This work is also in part supported by NSF grant IIS-2452129. We acknowledge the support by resources provided by the Office of Research Computing at George Mason University (https://orc.gmu.edu) and funded in part by grants from the NSF (Award Number 2018631). Any opinions, findings, and conclusions or recommendations expressed in this material are solely those of the authors.

\FloatBarrier
\bibliography{anthology,custom}

\appendix
\clearpage
\FloatBarrier
\section*{Appendix}
\label{sec:appendix}

Appendix~A.1 records implementation details, Appendix~A.2 gives the probe lists and translation checks, and Appendix~A.3 records the data and evaluation split. Appendix~A.4 organizes the additional results by the question each analysis answers.

\renewcommand{\thesection}{A.\arabic{section}}
\section{Complete Implementation Details}
\label{sec:appendix-implementation}
All runs use one Llama-style decoder with pre-norm RMSNorm and approximately 310M parameters. It has 12 layers, hidden size 768, FFN size 3072, 12 attention heads, and untied token/output matrices. Every checkpoint uses the unchanged \texttt{meta-llama/Llama-3.2-1B} BPE tokenizer and its complete 128{,}256-token multilingual vocabulary; we train one shared tokenizer and retain its full vocabulary in every condition. Section~\ref{sec:appendix-robustness} reports the tokenizer-identity audit. Throughput is fixed at $32{,}768$ tokens per step, using batch size $8$, gradient accumulation $8$, and sequence length $512$. For non-English probe construction, we use NLLB for translation, COMETKiwi for quality scoring, back-translation, and manual review of low-confidence cases.

\section{Probe Details}
\label{sec:probe-details}

\paragraph{Semantic Axes}
Table~\ref{tab:appendix-axes} lists every axis, and Table~\ref{tab:appendix-axis-grounding} gives the literature that motivated each theme. The endpoint pairs are hand-authored rather than taken verbatim from the cited sources, and they were fixed before analysis.

\begin{table*}[!tbp]
\centering
\footnotesize
\setlength{\tabcolsep}{1.5pt}
\input{tables/tab1-appendix-axes.tex}
\caption{Complete list of 50 semantic axes. The same fixed contrasts are reused in every language setting, and no endpoint pair is fitted to the reported results.}
\label{tab:appendix-axes}
\end{table*}

\begin{table*}[!tbp]
\centering
\footnotesize
\input{tables/tab2-appendix-axis-grounding-part1.tex}
\caption{Theme-level sources for all 50 semantic axes. The cited works motivate broad domains, not the exact hand-authored endpoint pairs.}
\label{tab:appendix-axis-grounding}
\end{table*}

\paragraph{Probe Translation and QC Summary}
Table~\ref{tab:appendix-probe-qc} reports translation checks by language. The corresponding result by quality tier appears in Section~\ref{sec:extra-results}.

\begin{table*}[!tbp]
\centering
\footnotesize
\input{tables/tab4-appendix-probe-qc.tex}
\caption{Translation checks for probes in all non-English languages. QE uses COMETKiwi when available (high $\geq$ 0.80, medium [0.60, 0.80), low $<$ 0.60); the table also reports manual-review flags and duplicate translations.}
\label{tab:appendix-probe-qc}
\end{table*}

\section{Data Statistics and Evaluation Split}
\label{sec:data-stats-splits}
Table~\ref{tab:appendix-data-stats-splits} records the data, model, and evaluation split used throughout the study.

\begin{center}
\centering
\footnotesize
\begin{tabularx}{\columnwidth}{@{}>{\raggedright\arraybackslash}p{0.39\columnwidth}>{\raggedright\arraybackslash}X@{}}
\toprule
Item & Statistics / split detail \\
\midrule
Pretraining data source & BabyBabelLM English and 8 Tier-1 non-English languages used in this study. \\
Target-language inventory & 8 languages spanning \LangSet. \\
Model configuration & Llama-style pre-norm decoder with RMSNorm, 12 layers, hidden size 768, FFN size 3072, 12 attention heads, vocabulary size 128{,}256, and untied token-embedding/output matrices (\(\sim\)310M parameters). \\
Token budget checkpoints & 1500 steps ($\approx$49.15M tokens) and 3000 steps ($\approx$98.30M tokens), at 32,768 tokens/step. \\
\ENtag{}-centered comparisons & 16 primary pairs (2 \ENONLY{} reference lengths $\times$ 8 \ENLtwo{} conditions); shared/separate-document conditions give 32 model-pair observations. Token-embedding/contextual-state results are two views of these pairs; English-only seed variation uses six ordered comparisons (three unique seed pairs) per checkpoint. \\
General-vocabulary anchors (alignment set) & 3000 English words; used only to estimate alignment maps. \\
Semantic probes (evaluation set) & 1000 terms; held out from alignment and used only to measure model differences. \\
Semantic axes & 50 axis pairs; fixed before training/evaluation and used only in evaluation metrics. \\
Negative-control set & 100 common physical-vocabulary words used only as an evaluation check. \\
Evaluation split & This unsupervised pretraining study has no supervised train/dev/test split. General-vocabulary words are used for alignment; separate semantic probes, axes, and negative controls are used only for evaluation. \\
\bottomrule
\end{tabularx}
\captionsetup{hypcap=false}
\captionof{table}{Data statistics and evaluation split. Words used for alignment are separate from the words used for evaluation.}
\label{tab:appendix-data-stats-splits}
\end{center}

\paragraph{Corpus and language-mixing check.}
BabyBabelLM provides language-specific mixtures of transcribed child--caregiver and adult conversations, educational material, children's books, child-oriented Wikipedia and news, and child-suitable film/television subtitles; where needed, it pads Tier-1 budgets with filtered OpenSubtitles and, for some languages, FineWeb-C or Wikipedia. All training data come from these BabyBabelLM source pools. Category proportions vary with language-specific source availability, so cross-language magnitudes remain descriptive. Within each target-language experiment, all conditions use the same data preparation procedure, and shared/separate conditions partition the same English pool.

Following \citet{shao-etal-2026-role}, we also ran a heuristic chunk-level fastText audit over all 3{,}213{,}686 documents used here. We flag a mixed-language candidate when the target and another language each cover at least 20\% of high-confidence sampled segments, with at least three such segments. This identifies 12{,}382 candidates (0.385\%). Among the 3{,}075{,}976 non-English documents, 2{,}721 (0.088\%) are target-plus-English candidates; per-language rates range from 0.001\% to 1.217\%. \citet{shao-etal-2026-role} found that bilingual documents comprised about 2\% of the 240B-token FineWeb mixtures they studied, using a different corpus and filter. We treat sparse mixed-language exposure as one candidate mechanism for future intervention studies.

\paragraph{Fixed training recipe and reported values.}
One pre-specified training recipe is reused for every run, keeping tuning choices fixed across conditions. All runs use the same 310M-parameter model configuration reported in Section~\ref{sec:training}, batch size $8$, gradient accumulation $8$, sequence length $512$, and throughput $32{,}768$ tokens/step, with checkpoints at $1500$ and $3000$ steps.

\section{Additional Results}
\label{sec:extra-results}

\subsection{Translation-Quality Pattern}
\begin{figure}[!tbp]
\centering
\includegraphics[width=\columnwidth]{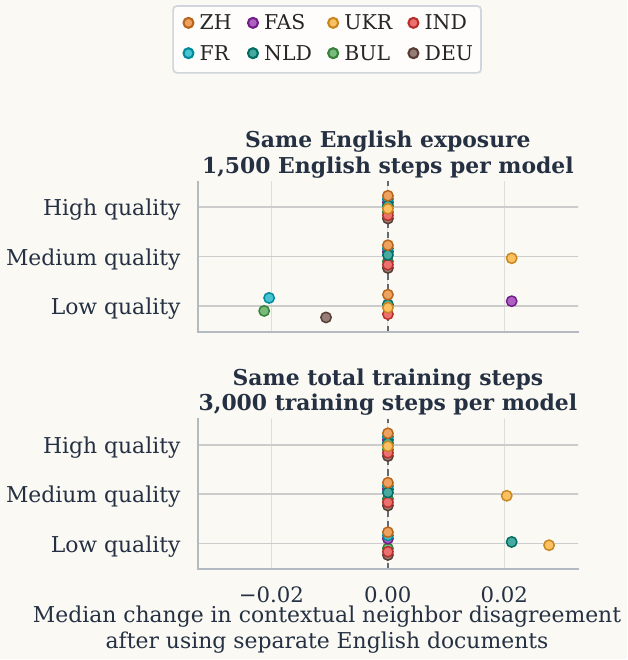}
\caption{\textbf{The document-overlap pattern is stable across translation-quality tiers.} Each point is one additional language. The upper panel holds English exposure fixed at 1{,}500 English training steps per model; the lower panel holds total training fixed at 3{,}000 steps per model. Values near zero indicate similar contextual neighbor disagreement after replacing shared English documents with separate sets.}
\label{fig:appendix-exp2-quality}
\end{figure}

Replacing shared English documents with separate sets changes contextual neighbor disagreement by similar amounts across translation-quality tiers. Figure~\ref{fig:appendix-exp2-quality} shows the pattern for all eight languages and both training controls.

\subsection{How to Read the Robustness Evidence}
\begin{figure*}[!tbp]
\centering
\includegraphics[width=0.94\textwidth]{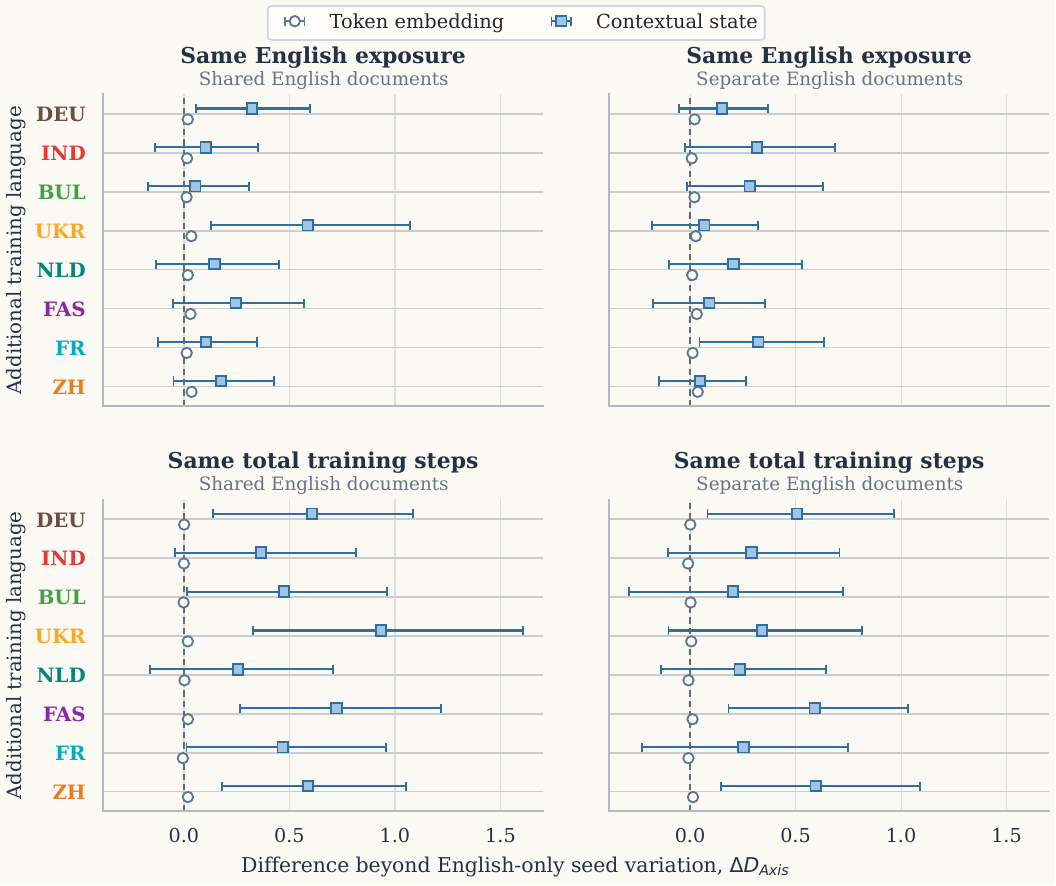}
\caption{\textbf{Language-level uncertainty under all four training controls.} Circles are token embeddings and squares are contextual states; horizontal bars are 95\% bootstrap intervals. The top row holds English exposure fixed, so both models receive 1{,}500 English training steps. The bottom row holds total training fixed at 3{,}000 steps. The right column replaces shared English documents with non-overlapping sets. Values are differences beyond the English-only seed mean.}
\label{fig:appendix-ci-guide}
\end{figure*}

The following results are organized by the question a reader may want to check. Table~\ref{tab:appendix-evidence-guide} points to the most informative display for each claim. Figure~\ref{fig:appendix-ci-guide} then shows all language-level intervals for the primary metric and all four training controls.

\begin{table*}[!tbp]
\centering
\footnotesize
\setlength{\tabcolsep}{4pt}
\renewcommand{\arraystretch}{1.12}
\begin{tabularx}{\textwidth}{@{}>{\raggedright\arraybackslash}p{0.22\textwidth}>{\raggedright\arraybackslash}p{0.25\textwidth}>{\raggedright\arraybackslash}X@{}}
\toprule
Question & Best evidence & Pattern to look for \\
\midrule
Does the difference exceed ordinary training variation? & Figure~\ref{fig:appendix-ci-guide} and Table~\ref{tab:same-language-controls} & Contextual-state points usually lie farther above the English-only seed reference than token-embedding points. \\
\midrule
Could English exposure, training-step count, or repeated documents explain it? & Table~\ref{tab:control-matrix}, Figures~\ref{fig:exp1-controls} and~\ref{fig:exp2-overlap} & The contextual difference remains under matched exposure, matched total training, and separate English documents. \\
\midrule
Is the result a consequence of the fitted coordinate map? & Table~\ref{tab:exp5-alignment}, Figure~\ref{fig:exp5-alignment}, and Table~\ref{tab:anchor-sensitivity} & Contextual-state differences remain after contextual-anchor and affine alignment and after changing the anchor subset. \\
\midrule
When does the difference appear? & Figures~\ref{fig:layer-norm-panel} and~\ref{fig:dense-progress} & It is visible at the earliest saved checkpoint, becomes largest in middle layers, and survives vector-length controls. \\
\midrule
Is it confined to the semantic probe categories? & Table~\ref{tab:exp1-negctrl} and Figure~\ref{fig:appendix-category-heatmap} & Physical-vocabulary controls also differ, while category magnitudes vary. The result is broader than a uniquely cultural effect. \\
\bottomrule
\end{tabularx}
\caption{\textbf{A reader's guide to the robustness evidence.} Each row links a possible alternative explanation to the display that tests it and states the visual pattern needed to evaluate the claim.}
\label{tab:appendix-evidence-guide}
\end{table*}

\begin{table*}[!tbp]
\centering
\footnotesize
\input{tables/tab9-exp1-negative-controls.tex}
\caption{\textbf{Physical-vocabulary controls show that the difference extends beyond the semantic probe categories.} The 100 controls are common physical-vocabulary terms. Their seed-centered disagreement is comparable to or larger than the primary probe summary, which places the result at the level of general representation change rather than a uniquely cultural effect.}
\label{tab:exp1-negctrl}
\end{table*}

\subsection{Matched Total Training and Same-Language Controls}
\label{sec:appendix-control-companions}
The main figures emphasize the setting in which the English-only and bilingual models receive the same amount of English. The following displays isolate the complementary setting in which both models instead receive 3{,}000 total training steps. Here the English-only model receives all steps in English, while the bilingual model receives 1{,}500 English steps and 1{,}500 steps in the additional language.

\begin{figure*}[!tbp]
\centering
\includegraphics[width=0.85\textwidth]{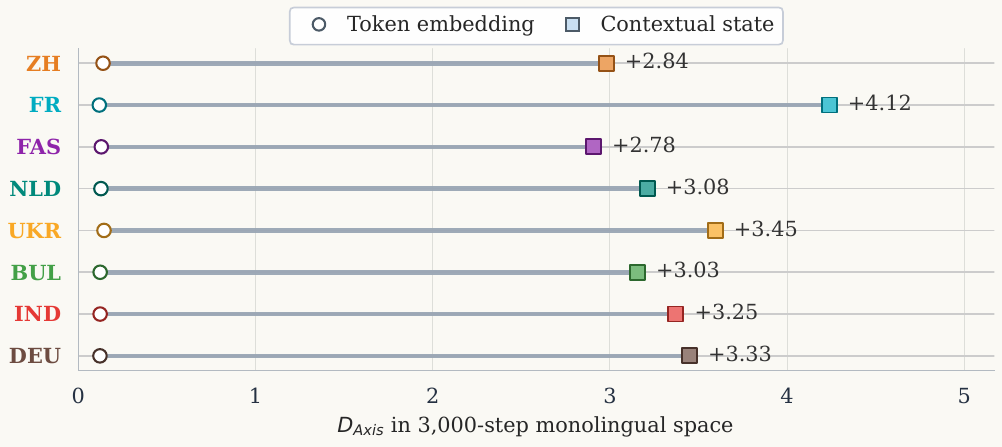}
\caption{\textbf{The representation-level ordering also holds in the additional-language spaces under matched total training.} Each monolingual reference receives 3{,}000 steps in the additional language, while its paired bilingual model receives 1{,}500 steps in English and 1{,}500 in that language. Contextual states differ more than token embeddings for all eight languages.}
\label{fig:appendix-exp3-multilingual-100m}
\end{figure*}

\begin{figure*}[!tbp]
\centering
\includegraphics[width=0.85\textwidth]{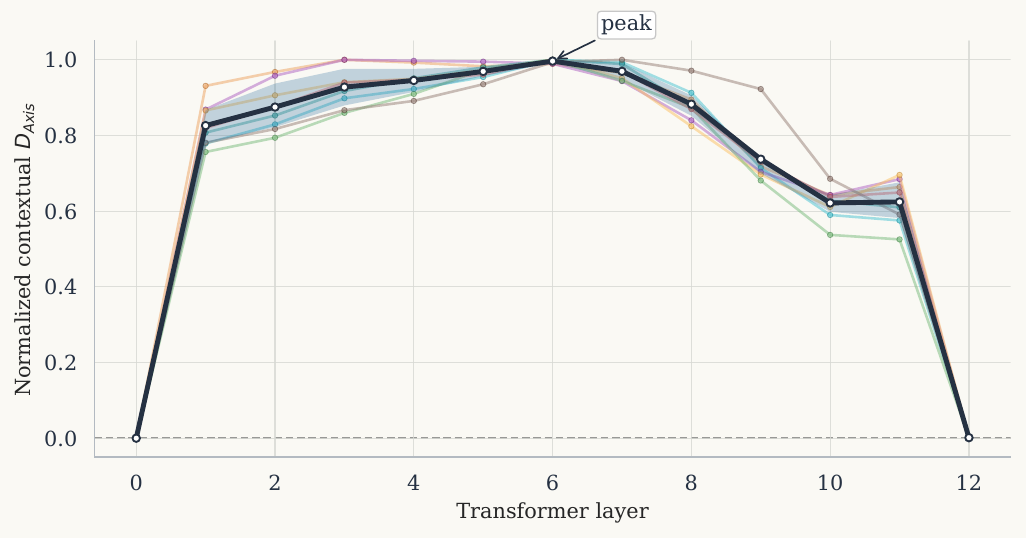}
\caption{\textbf{The middle-layer peak remains when total training is matched.} The English-only and bilingual models each receive 3{,}000 training steps. Thin lines show the eight additional-language conditions, and the dark line is their mean.}
\label{fig:appendix-exp3-layerwise-100m}
\end{figure*}

\begin{figure*}[!tbp]
\centering
\includegraphics[width=0.85\textwidth]{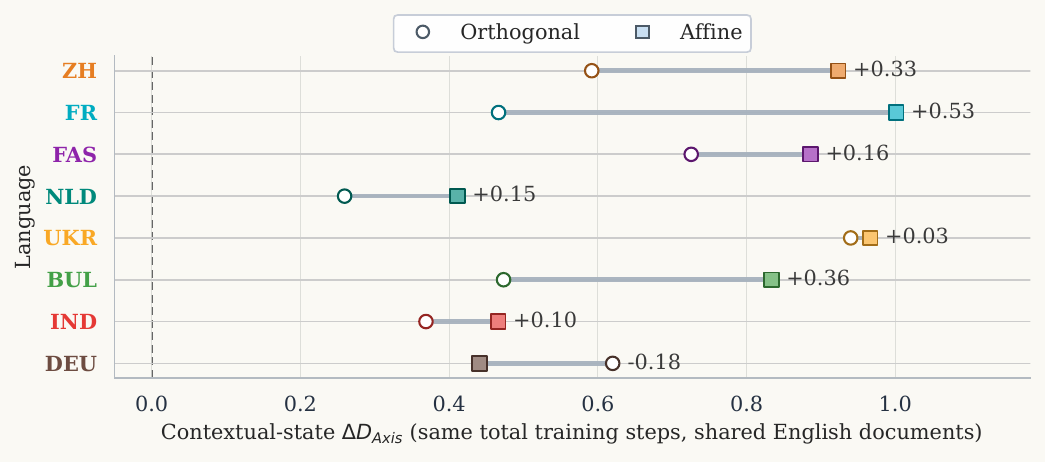}
\caption{\textbf{Alternative maps preserve the matched-total-training result.} Lines connect the orthogonal and affine contextual-state values for each additional training language when both models receive 3{,}000 training steps. The map changes the numerical scale but does not reduce the result to the English-only seed reference.}
\label{fig:appendix-exp4-align-100m}
\end{figure*}

\paragraph{Same-language controls.}
Six ordered English-only comparisons, corresponding to three unique seed pairs, define the seed-variation mean used in the English-centered analyses.
\begin{table*}[!tbp]
\centering
\footnotesize
\input{tables/tab25-same-language-controls.tex}
\caption{\textbf{English-only runs define the reference for ordinary training variation.} Raw \(D_{Axis}\) is summarized across six ordered comparisons, corresponding to three unique seed pairs, for each training length and representation. The matching mean is subtracted from each bilingual comparison reported as \(\Delta D_{Axis}\).}
\label{tab:same-language-controls}
\end{table*}

\paragraph{Axis-group holdout.}
This analysis tests whether one axis group drives the contextual difference. Both matched and held-out groups retain positive seed-centered values.
\begin{table*}[!tbp]
\centering
\footnotesize
\input{tables/tab26-framework-holdout.tex}
\caption{\textbf{The contextual result transfers to held-out axis groups.} The axes used for the matched column and those reserved for the held-out column are disjoint. Contextual-state differences remain elevated for both groups, so one handcrafted theme does not drive the aggregate pattern.}
\label{tab:framework-holdout}
\end{table*}

\subsection{Robustness Beyond the Main Controls}
\label{sec:appendix-robustness}
\paragraph{Training-progress sensitivity.}
These results ask whether the difference between contextual states and token embeddings appears only late in training. Across all eight language conditions, it is present at the earliest saved checkpoint and persists throughout training.

\begin{figure*}[!tbp]
\centering
\includegraphics[width=0.96\textwidth]{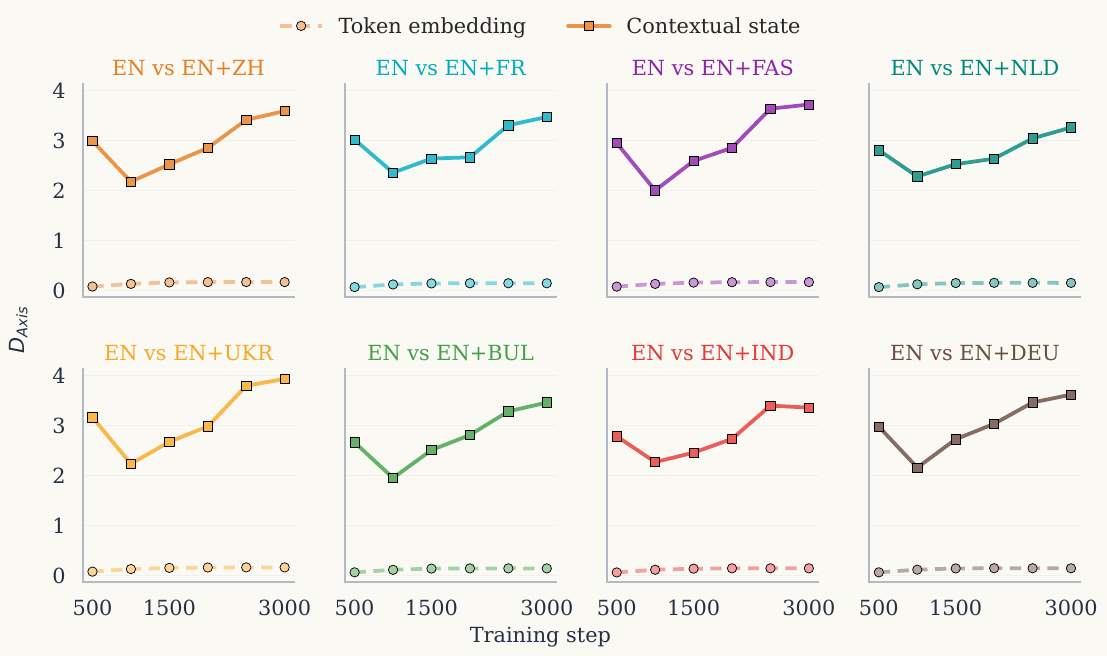}
\caption{\textbf{Contextual states differ more than token embeddings throughout training.} Each panel traces one additional-language condition at 500-step intervals when both models see 1{,}500 English training steps and reuse the same English documents. The difference is present at the earliest saved checkpoint and remains through 3{,}000 steps.}
\label{fig:dense-progress}
\end{figure*}

\paragraph{Alternative alignment maps.}
Switching to an affine map changes the values but does not bring contextual \(D_{Axis}\) close to English-only seed variation.
\begin{figure*}[!tbp]
\centering
\includegraphics[width=0.85\textwidth]{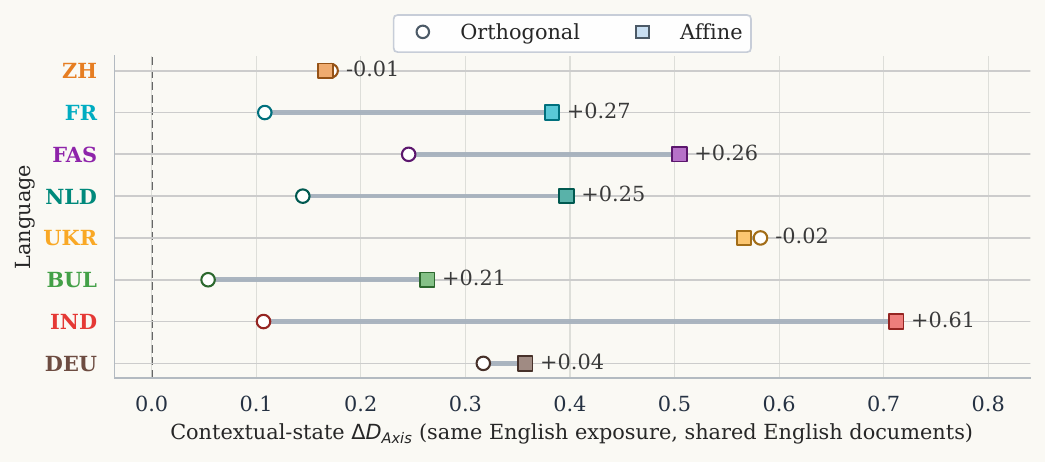}
\caption{\textbf{The alignment method changes values, not the main result.} Lines connect orthogonal and affine contextual-state values for each additional training language when both models see 1{,}500 English training steps and share the English documents.}
\label{fig:exp5-alignment}
\end{figure*}

\paragraph{Anchor words.}
The contextual-state difference is stable when the alignment map is fitted using random subsets of 500 to 3{,}000 English anchor words.
\begin{table*}[!tbp]
\centering
\footnotesize
\setlength{\tabcolsep}{6pt}
\input{tables/tab28-anchor-sensitivity.tex}
\caption{\textbf{The representation-level pattern is stable as the alignment set grows from 500 to 3{,}000 words.} Held-out \(D_{Axis}\) is averaged over eight additional languages and five random anchor draws. A single value means the rounded result is unchanged across subset sizes. The arrows show that alignment fit improves with more anchors while contextual-state differences remain much larger than token-embedding differences.}
\label{tab:anchor-sensitivity}
\end{table*}

\paragraph{Nearest-neighbor parameter sensitivity.}
Table~\ref{tab:knn-sensitivity} varies the neighborhood size used by the complementary \(D_{NN}\) metric. Mean centered disagreement remains positive across all tested values of \(k\), although the ordering of embeddings and contextual states varies with \(k\). Our main comparison therefore rests on the primary \(D_{Axis}\) metric rather than on one neighborhood size.
\begin{table*}[!tbp]
\centering
\footnotesize
\setlength{\tabcolsep}{4pt}
\input{tables/tab32-knn-sensitivity.tex}
\caption{\textbf{The complementary neighbor metric remains above seed variation across neighborhood sizes.} English-only values summarize the six ordered same-language comparisons; bilingual values summarize the sixteen shared-document comparisons across the matched-exposure and matched-total-training settings. The last column in each representation block is their difference.}
\label{tab:knn-sensitivity}
\end{table*}

\paragraph{Tokenizer identity.}
SHA-256 hashes are identical across all runs for the tokenizer vocabulary, configuration, and special-token map. A condition-specific tokenizer therefore cannot explain the result.

\begin{figure*}[!tbp]
\centering
\includegraphics[width=0.95\textwidth]{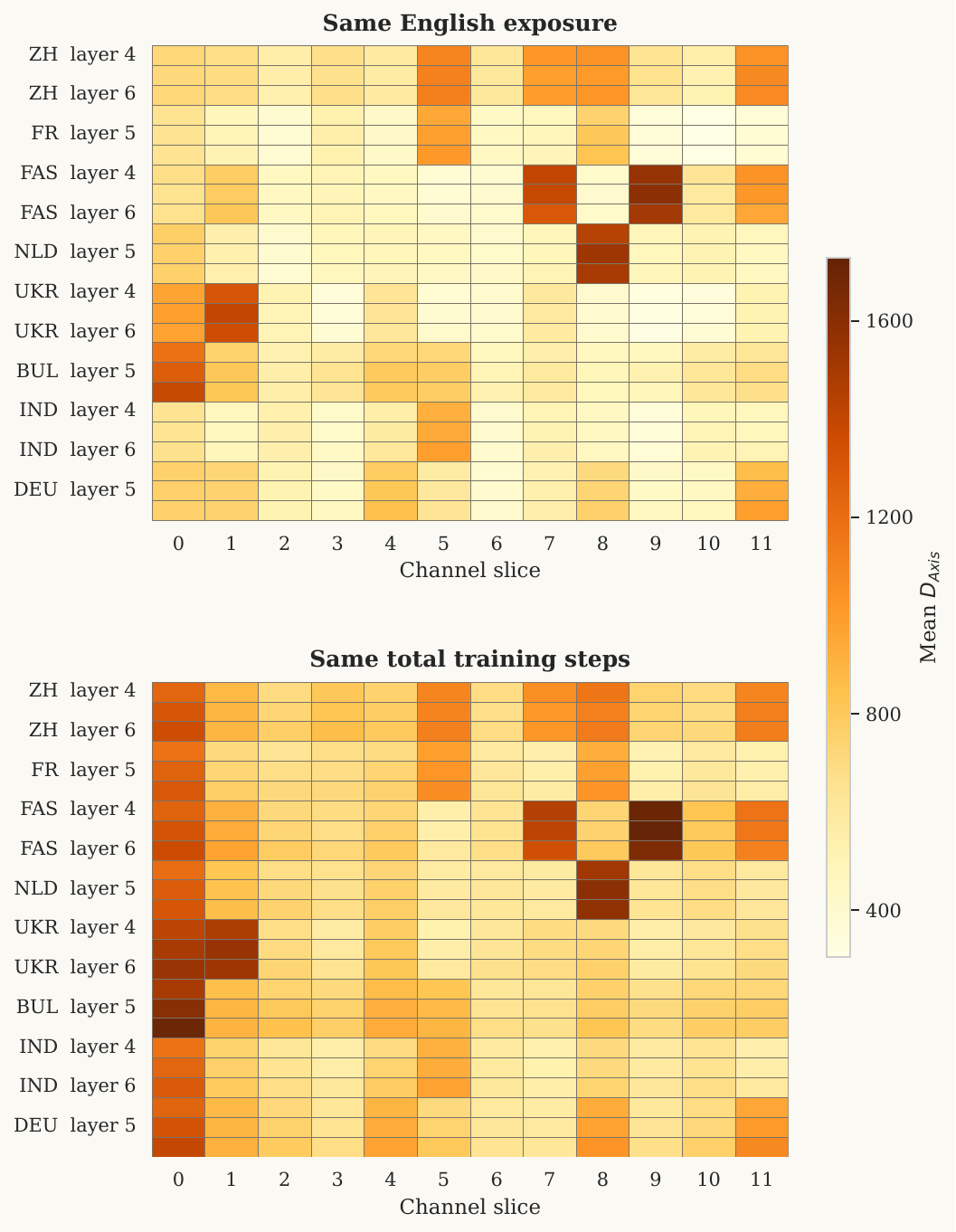}
\caption{\textbf{The middle-layer difference is distributed across channel slices.} Each cell reports mean $D_{Axis}$ for one contiguous head-width slice of the hidden dimension.}
\label{fig:appendix-perhead}
\end{figure*}

\paragraph{Channel-slice diagnostic.}
We divide the hidden dimension into contiguous head-width channel slices and compute the metric within each slice. Figure~\ref{fig:appendix-perhead} shows a distributed middle-layer pattern rather than one isolated channel group. Attention-head output interventions offer a complementary direction for future work.

\subsection{Differences Across Categories and Languages}
\label{sec:appendix-multilingual}
\paragraph{Category and word results.}
The following analyses show which probes contribute most to the contextual-state difference, with signed examples and breakdowns by category and word.

\begin{figure*}[p]
\centering
\begin{minipage}[t]{\textwidth}
\centering
\footnotesize
\setlength{\tabcolsep}{6pt}
\input{tables/tab30-aggregate-scope-tests.tex}
\captionof{table}{\textbf{Aggregate tests confirm the representation-level gap.} The two training settings state which quantity is held fixed.}
\label{tab:aggregate-tests}
\end{minipage}

\vspace{0.6em}

\begin{minipage}[t]{\textwidth}
\centering
\footnotesize
\input{tables/tab20-exp1-hotspots.tex}
\captionof{table}{\textbf{Contextual neighbor disagreement varies substantially across probes.}}
\label{tab:appendix-hotspots}
\end{minipage}

\vspace{0.6em}

\includegraphics[width=0.80\textwidth]{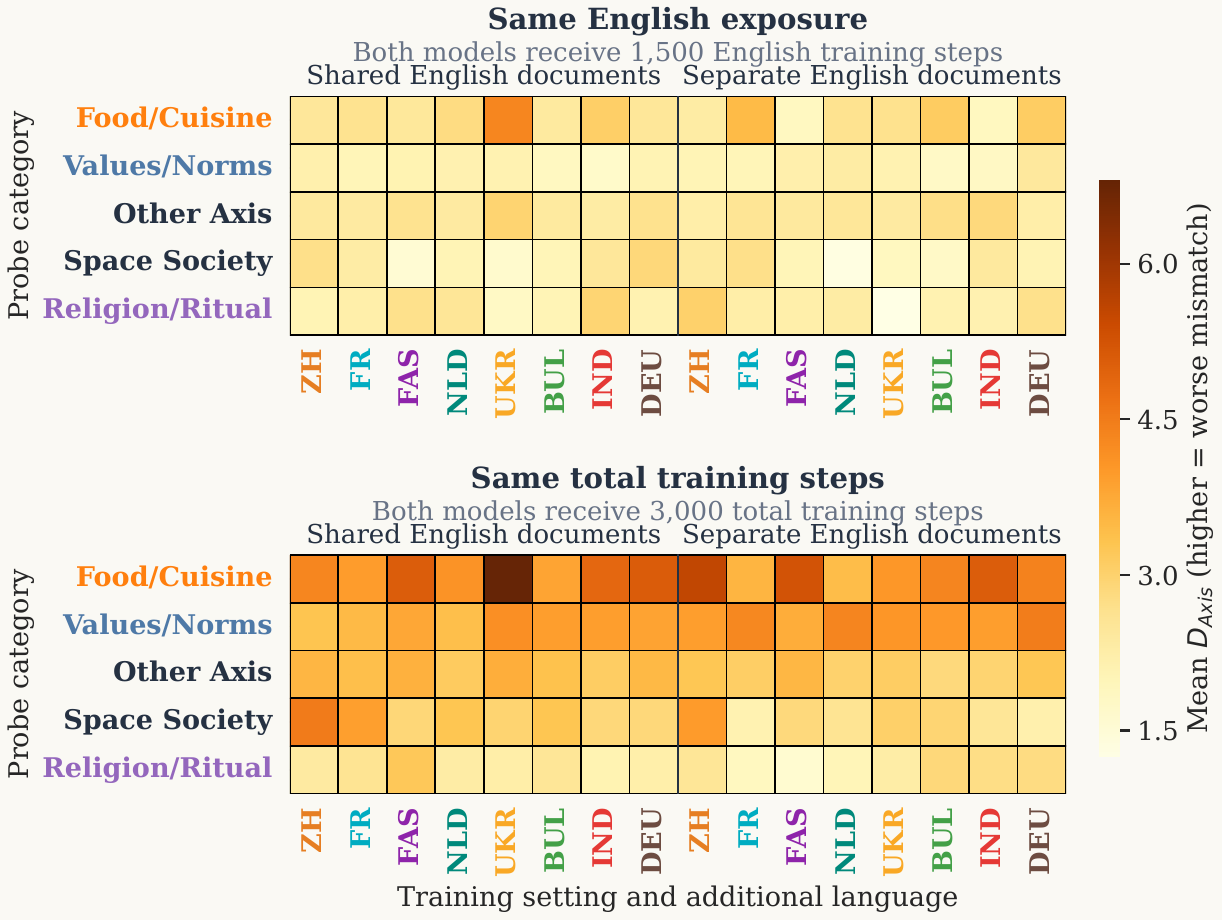}
\caption{\textbf{Contextual differences vary across probe categories.} \DomFood{}, \DomSymbols{}, and \DomIdentity{} contribute disproportionately, while the physical-vocabulary controls in Table~\ref{tab:exp1-negctrl} place the effect at the level of general representation change.}
\label{fig:appendix-category-heatmap}
\end{figure*}

\begin{figure*}[p]
\centering
\includegraphics[width=0.92\textwidth,height=0.80\textheight,keepaspectratio]{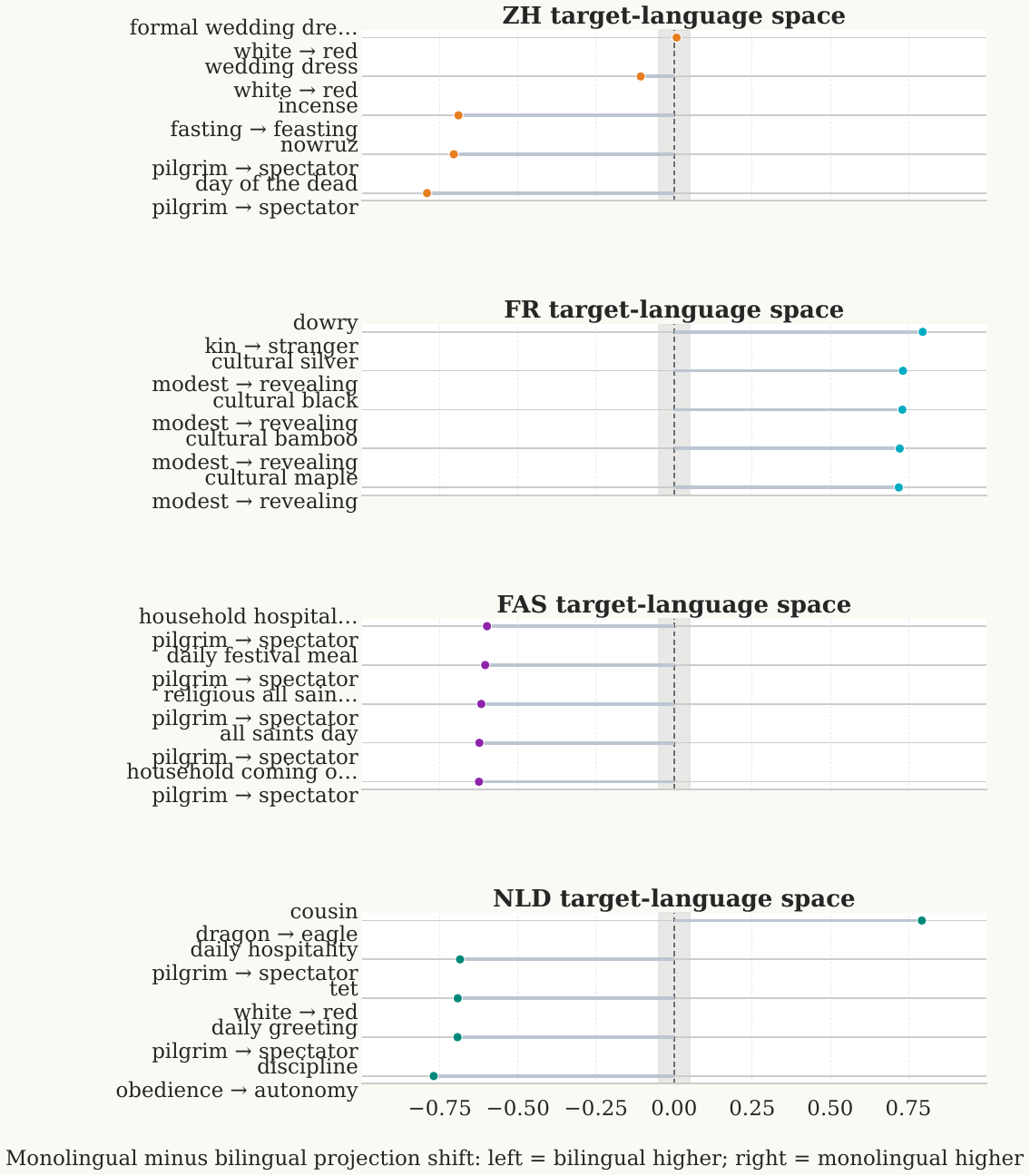}
\caption{\textbf{Largest signed shifts in the Chinese, French, Persian, and Dutch target-language spaces.} Each panel names the probe and semantic contrast and reports the monolingual-minus-bilingual projection difference.}
\label{fig:appendix-l2-signed-hotspots}
\end{figure*}

\begin{figure*}[p]
\centering
\includegraphics[width=0.92\textwidth,height=0.80\textheight,keepaspectratio]{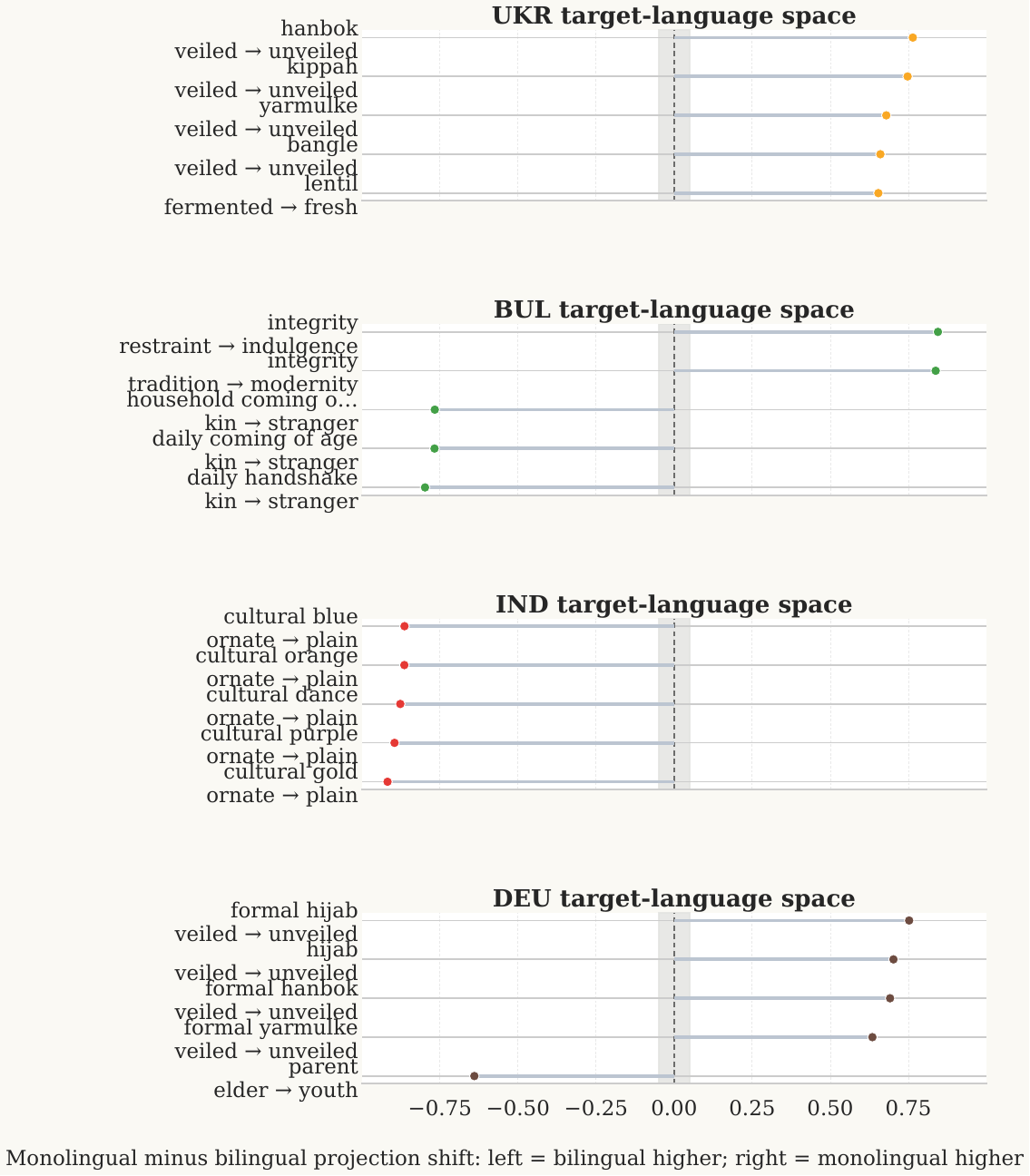}
\caption{\textbf{Largest signed shifts in the Ukrainian, Bulgarian, Indonesian, and German target-language spaces.} Each panel names the probe and semantic contrast and reports the monolingual-minus-bilingual projection difference.}
\label{fig:appendix-l2-signed-hotspots-continued}
\end{figure*}

\begin{figure*}[!tbp]
\centering
\begin{subfigure}[t]{0.98\textwidth}
\centering
\includegraphics[width=0.98\linewidth]{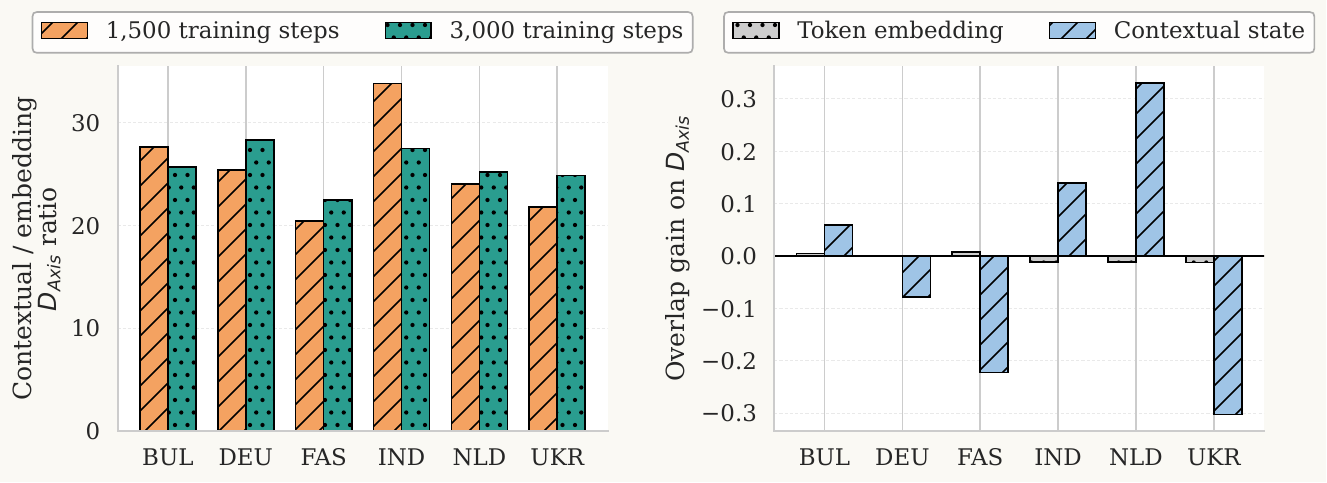}
\caption{Target-space representation ratios and the change after replacing shared English documents with separate sets. The two monolingual references receive 1{,}500 and 3{,}000 training steps.}
\label{fig:appendix-multilingual-overview}
\end{subfigure}
\begin{subfigure}[t]{0.88\textwidth}
\centering
\includegraphics[width=0.96\linewidth]{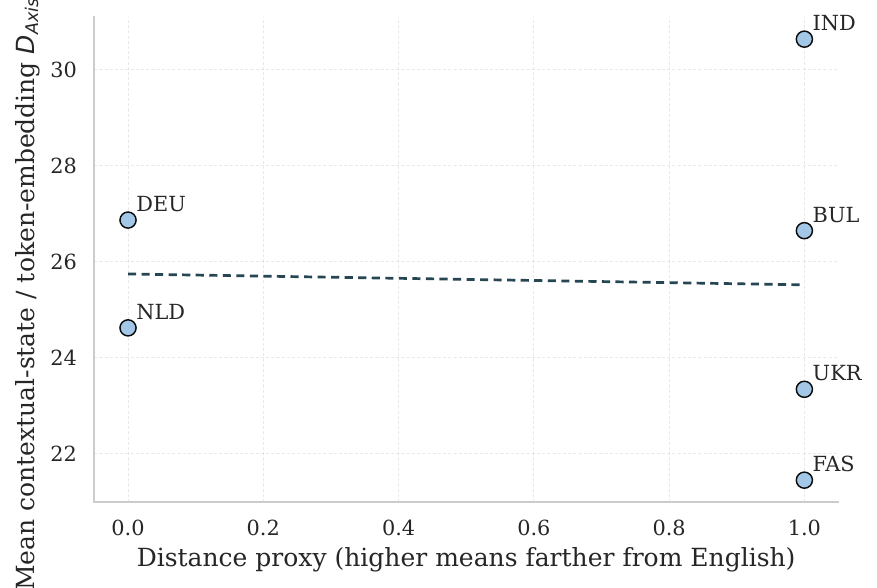}
\caption{Exploratory comparison of six languages using coarse script and family indicators; the points show no monotonic relation.}
\label{fig:appendix-typology}
\end{subfigure}
\caption{\textbf{Per-language patterns for six additional conditions preserve the representation-level ordering.} Contextual-state differences remain larger than token-embedding differences, while the exploratory typology comparison shows no clear monotonic relationship.}
\label{fig:appendix-multilingual-summary}
\end{figure*}

\paragraph{Additional languages.}
Figure~\ref{fig:appendix-multilingual-summary} completes the per-language view for the six languages outside the earlier Chinese and French analysis. Panel~(a) compares raw target-language differences in contextual states and token embeddings, then shows how separating English documents changes each metric. Panel~(b) places the six languages against coarse script and family indicators; the plotted points show no clear monotonic relationship.

\end{document}

%% file: tables/tab17-exp4-alignment-methods.tex
\begin{tabular*}{\textwidth}{@{\extracolsep{\fill}}lllrrrr@{}}
\toprule
& & & \multicolumn{3}{c}{Held-out difference (mean $\pm$ SD)} & \multicolumn{1}{c}{Alignment fit} \\
\cmidrule(lr){4-6}\cmidrule(l){7-7}
Representation & Alignment & Fitted on & $D_{NN}$ & $D_{Struct}$ & $D_{Axis}$ & Error / word \\
\midrule
Token embedding & Orthogonal & Token embedding & $0.029 \pm 0.019$ & $0.014 \pm 0.015$ & $0.014 \pm 0.012$ & 0.041 \\
 & Orthogonal & Contextual state & $0.029 \pm 0.019$ & $0.014 \pm 0.015$ & $0.014 \pm 0.012$ & 0.194 \\
 & Affine & Token embedding & $0.097 \pm 0.026$ & $0.055 \pm 0.024$ & $0.021 \pm 0.023$ & 0.030 \\
\cmidrule(lr){1-7}
Contextual state & Orthogonal & Token embedding & $0.015 \pm 0.016$ & $0.042 \pm 0.038$ & $0.386 \pm 0.248$ & 0.041 \\
 & Orthogonal & Contextual state & $0.015 \pm 0.016$ & $0.042 \pm 0.038$ & $0.386 \pm 0.248$ & 0.194 \\
 & Affine & Token embedding & $0.027 \pm 0.016$ & $0.052 \pm 0.043$ & $0.580 \pm 0.260$ & 0.030 \\
\bottomrule
\end{tabular*}

%% file: tables/tab-main-axis-inventory.tex
\begin{tabular*}{\linewidth}{@{\extracolsep{\fill}}lccr@{}}
\toprule
Theme & Endpoint 1 & Endpoint 2 & Count \\
\midrule
Values and norms       & individualism & collectivism & 10 \\
Family and kinship     & mother        & father       & 5 \\
Religion and ritual    & sacred        & secular      & 5 \\
Governance and law     & democracy     & monarchy     & 5 \\
Food and cuisine       & rice          & bread        & 5 \\
Festivals and holidays & fasting       & feasting     & 5 \\
Clothing and appearance& veiled        & unveiled     & 5 \\
Symbols and colors     & white         & red          & 5 \\
Social identity        & local         & global       & 5 \\
\bottomrule
\end{tabular*}

%% file: tables/tab-main-translation-quality.tex
\begin{tabular*}{\linewidth}{@{\extracolsep{\fill}}lrrr@{}}
\toprule
Language & \begin{tabular}[c]{@{}r@{}}Mean / $\geq 0.80$\\score (\%)\end{tabular} & \begin{tabular}[c]{@{}r@{}}Review /\\duplicate (\%)\end{tabular} & \begin{tabular}[c]{@{}r@{}}Largest median\\overlap change\end{tabular} \\
\midrule
BUL & 0.797 / 71.7 & 37.1 / 4.7  & 0.021 \\
DEU & 0.779 / 66.6 & 35.1 / 4.7  & 0.011 \\
FAS & 0.797 / 69.6 & 51.0 / 11.8 & 0.021 \\
FR  & 0.777 / 65.8 & 28.8 / 3.2  & 0.020 \\
IND & 0.764 / 59.9 & 38.9 / 9.0  & 0.000 \\
NLD & 0.758 / 58.3 & 35.1 / 5.5  & 0.021 \\
UKR & 0.791 / 69.5 & 32.8 / 6.0  & 0.028 \\
ZH  & 0.745 / 52.9 & 49.1 / 13.1 & 0.000 \\
\bottomrule
\end{tabular*}

%% file: tables/tab1-appendix-axes.tex
\begin{tabular}{@{}rll@{\hspace{1.2em}}rll@{}}
\toprule
\multicolumn{3}{c}{Axis set A} & \multicolumn{3}{c}{Axis set B} \\
\cmidrule(lr){1-3}\cmidrule(lr){4-6}
\# & Endpoint 1 & Endpoint 2 & \# & Endpoint 1 & Endpoint 2 \\
\midrule
1 & individualism & collectivism & 26 & rice & bread \\
2 & hierarchy & equality & 27 & tea & coffee \\
3 & duty & freedom & 28 & spicy & mild \\
4 & honor & shame & 29 & vegetarian & meat \\
5 & obedience & autonomy & 30 & fermented & fresh \\
6 & tradition & modernity & 31 & fasting & feasting \\
7 & restraint & indulgence & 32 & lunar & solar \\
8 & conformity & dissent & 33 & mourning & celebration \\
9 & solidarity & competition & 34 & ancestor & novelty \\
10 & certainty & ambiguity & 35 & pilgrim & spectator \\
11 & mother & father & 36 & veiled & unveiled \\
12 & elder & youth & 37 & formal & casual \\
13 & kin & stranger & 38 & ornate & plain \\
14 & lineage & mobility & 39 & modest & revealing \\
15 & clan & individual & 40 & uniform & personalized \\
16 & sacred & secular & 41 & white & red \\
17 & ritual & routine & 42 & black & gold \\
18 & prayer & reason & 43 & dragon & eagle \\
19 & pilgrimage & tourism & 44 & lotus & rose \\
20 & taboo & permissive & 45 & script & image \\
21 & democracy & monarchy & 46 & local & global \\
22 & federal & centralized & 47 & indigenous & diaspora \\
23 & customary & codified & 48 & assimilation & pluralism \\
24 & authority & deliberation & 49 & majority & minority \\
25 & patronage & meritocracy & 50 & homeland & migration \\
\bottomrule
\end{tabular}

%% file: tables/tab2-appendix-axis-grounding-part1.tex
\begin{tabular}{@{}>{\raggedleft\arraybackslash}p{0.04\textwidth}>{\raggedright\arraybackslash}p{0.15\textwidth}>{\raggedright\arraybackslash}p{0.15\textwidth}>{\raggedright\arraybackslash}p{0.25\textwidth}>{\raggedright\arraybackslash}p{0.28\textwidth}@{}}
\toprule
\# & Endpoint 1 & Endpoint 2 & Category & Citation(s) \\
\midrule
1 & individualism & collectivism & Values and social norms & \cite{hofstede2001culture,schwartz2006theory} \\
2 & hierarchy & equality & Values and social norms & \cite{hofstede2001culture,schwartz2006theory} \\
3 & duty & freedom & Values and social norms & \cite{inglehart2005modernization} \\
4 & honor & shame & Values and social norms & \cite{house2004culture} \\
5 & obedience & autonomy & Values and social norms & \cite{schwartz2006theory} \\
6 & tradition & modernity & Values and social norms & \cite{inglehart2005modernization} \\
7 & restraint & indulgence & Values and social norms & \cite{hofstede2001culture} \\
8 & conformity & dissent & Values and social norms & \cite{schwartz2006theory} \\
9 & solidarity & competition & Values and social norms & \cite{house2004culture} \\
10 & certainty & ambiguity & Values and social norms & \cite{hofstede2001culture} \\
11 & mother & father & Family and kinship & \cite{house2004culture} \\
12 & elder & youth & Family and kinship & \cite{hofstede2001culture} \\
13 & kin & stranger & Family and kinship & \cite{schwartz2006theory} \\
14 & lineage & mobility & Family and kinship & \cite{inglehart2005modernization} \\
15 & clan & individual & Family and kinship & \cite{schwartz2006theory} \\
16 & sacred & secular & Religion and ritual life & \cite{inglehart2005modernization} \\
17 & ritual & routine & Religion and ritual life & \cite{house2004culture} \\
18 & prayer & reason & Religion and ritual life & \cite{inglehart2005modernization} \\
19 & pilgrimage & tourism & Religion and ritual life & \cite{hershcovich2022challenges} \\
20 & taboo & permissive & Religion and ritual life & \cite{hofstede2001culture} \\
21 & democracy & monarchy & Governance, institutions, and law & \cite{hofstede2001culture} \\
22 & federal & centralized & Governance, institutions, and law & \cite{house2004culture} \\
23 & customary & codified & Governance, institutions, and law & \cite{hofstede2001culture} \\
24 & authority & deliberation & Governance, institutions, and law & \cite{house2004culture} \\
25 & patronage & meritocracy & Governance, institutions, and law & \cite{hofstede2001culture} \\
\bottomrule
\end{tabular}

%% file: tables/tab4-appendix-probe-qc.tex
\begin{tabular}{@{}lrrrcccrr@{}}
\toprule
Lang & $n$ & Mean QE & Median QE & High & Medium & Low & Manual-review & Duplicate-target \\
\midrule
BUL & 1000 & 0.796 & 0.847 & 717 & 192 & 91 & 371 & 47 \\
DEU & 1000 & 0.779 & 0.824 & 666 & 242 & 92 & 351 & 47 \\
FAS & 1000 & 0.797 & 0.833 & 696 & 249 & 55 & 510 & 118 \\
FR & 1000 & 0.777 & 0.832 & 658 & 227 & 115 & 288 & 32 \\
IND & 1000 & 0.763 & 0.820 & 599 & 262 & 139 & 389 & 90 \\
NLD & 1000 & 0.758 & 0.820 & 583 & 270 & 147 & 351 & 55 \\
UKR & 1000 & 0.791 & 0.841 & 695 & 210 & 95 & 328 & 60 \\
ZH & 1000 & 0.745 & 0.806 & 529 & 312 & 159 & 491 & 131 \\
\bottomrule
\end{tabular}

%% file: tables/tab9-exp1-negative-controls.tex
\begin{tabular}{@{}llrrr@{}}
\toprule
Representation & Group & $\Delta D_{NN}$ & $\Delta D_{Struct}$ & $\Delta D_{Axis}$ \\
\midrule
Token embedding & Cultural probes & 0.029 & 0.014 & 0.014 \\
 & Negative controls & 0.328 & 0.017 & 0.075 \\
\cmidrule(lr){1-5}
Contextual state & Cultural probes & 0.015 & 0.042 & 0.386 \\
 & Negative controls & 0.165 & 0.034 & 0.483 \\
\bottomrule
\end{tabular}

%% file: tables/tab25-same-language-controls.tex
\begin{tabular}{@{}llrrr@{}}
\toprule
Setting & Representation & Mean raw $D_{Axis}$ & Std. dev. & English seed pairs \\
\midrule
Same English exposure & Contextual state & 2.258 & 0.182 & 6 \\
Same English exposure & Token embedding & 0.113 & 0.002 & 6 \\
Same total training steps & Contextual state & 3.204 & 0.295 & 6 \\
Same total training steps & Token embedding & 0.150 & 0.002 & 6 \\
\bottomrule
\end{tabular}

%% file: tables/tab26-framework-holdout.tex
\begin{tabular}{@{}lrrr@{}}
\toprule
Representation & Matched-axis $D_{Axis}$ & Held-out-axis $D_{Axis}$ & Held-out minus matched \\
\midrule
Contextual state & 0.261 & 0.390 & 0.129 \\
Token embedding & 0.018 & 0.014 & -0.004 \\
\bottomrule
\end{tabular}

%% file: tables/tab28-anchor-sensitivity.tex
\begin{tabular}{@{}lcccc@{}}
\toprule
 & \multicolumn{2}{c}{Held-out mean $D_{Axis}$} & \multicolumn{2}{c}{Alignment residual per anchor} \\
\cmidrule(lr){2-3}\cmidrule(l){4-5}
Setting & Token embedding & Contextual state & Token embedding & Contextual state \\
\midrule
Same English exposure & 0.135 & 2.439 & 0.069 $\rightarrow$ 0.038 & 0.378 $\rightarrow$ 0.170 \\
Same total training steps & 0.155 & 3.552 & 0.081 $\rightarrow$ 0.044 & 0.495 $\rightarrow$ 0.217 \\
\bottomrule
\end{tabular}

%% file: tables/tab32-knn-sensitivity.tex
\begin{tabular}{@{}rrrrrrr@{}}
\toprule
& \multicolumn{3}{c}{Token embedding} & \multicolumn{3}{c}{Contextual state} \\
\cmidrule(lr){2-4}\cmidrule(l){5-7}
$k$ & English-only & Bilingual & Difference & English-only & Bilingual & Difference \\
\midrule
5   & 0.568 & 0.590 & 0.023 & 0.807 & 0.827 & 0.019 \\
10  & 0.646 & 0.668 & 0.022 & 0.796 & 0.809 & 0.013 \\
25  & 0.613 & 0.642 & 0.029 & 0.772 & 0.787 & 0.015 \\
50  & 0.610 & 0.653 & 0.043 & 0.781 & 0.804 & 0.023 \\
100 & 0.779 & 0.803 & 0.024 & 0.778 & 0.803 & 0.025 \\
\bottomrule
\end{tabular}

%% file: tables/tab30-aggregate-scope-tests.tex
\begin{tabular}{@{}llrrr@{}}
\toprule
Test block & Slice & $n$ & Mean diff. & One-sided $p$ \\
\midrule
Main EN-centered gap & Same English exposure & 16 & 0.181 & 1.53e-05 \\
Main EN-centered gap & Same total training steps & 16 & 0.464 & 1.53e-05 \\
Main EN-centered gap & All & 32 & 0.322 & $< 10^{-6}$ \\
Anchor-subset reruns & Same English exposure & 160 & 2.304 & $< 10^{-6}$ \\
Anchor-subset reruns & Same total training steps & 160 & 3.397 & $< 10^{-6}$ \\
Anchor-subset reruns & All & 320 & 2.850 & $< 10^{-6}$ \\
\bottomrule
\end{tabular}

%% file: tables/tab20-exp1-hotspots.tex
\begin{tabular}{@{}r l r r l r@{}}
\toprule
\multicolumn{3}{c}{Highest contextual divergence} & \multicolumn{3}{c}{Lowest contextual divergence} \\
\cmidrule(lr){1-3}\cmidrule(lr){4-6}
Rank & Word & Mean $D_{NN}$ & Rank & Word & Mean $D_{NN}$ \\
\midrule
1 & vegan & 0.999 & 1 & extended mother & 0.323 \\
2 & stereotype & 0.997 & 2 & public democracy & 0.333 \\
3 & bamboo & 0.997 & 3 & extended father & 0.338 \\
4 & integrity & 0.996 & 4 & extended son & 0.341 \\
5 & carnival & 0.996 & 5 & extended daughter & 0.361 \\
6 & jewelry & 0.996 & 6 & sacred prayer & 0.362 \\
7 & espresso & 0.996 & 7 & public ministry & 0.373 \\
8 & lineage & 0.996 & 8 & sacred mosque & 0.373 \\
9 & homeland & 0.996 & 9 & local majority & 0.381 \\
10 & governance & 0.995 & 10 & sacred curse & 0.384 \\
\bottomrule
\end{tabular}